\documentclass{article}

\PassOptionsToPackage{numbers, compress}{natbib}

 \usepackage[preprint]{neurips_2024}

\usepackage[utf8]{inputenc} 
\usepackage[T1]{fontenc}    
\usepackage{hyperref}       
\usepackage{url}            
\usepackage{booktabs}       
\usepackage{amsfonts}       
\usepackage{nicefrac}       
\usepackage{microtype}      
\usepackage{xcolor}         
\newcommand{\mytextsf}[1]{{\textsf{#1}}}
\newcommand{\lexus}{\mytextsf{ASTRA}}
\newcommand{\ngame}{\mytextsf{NGAME}}
\newcommand{\renee}{\mytextsf{Ren\'ee}}
\newcommand{\dexa}{\mytextsf{DEXA}}

\newcommand{\elias}{\mytextsf{ELIAS}}

\newcommand{\cascadexml}{\mytextsf{CascadeXML}}
\newcommand{\attentionxml}{\mytextsf{AttentionXML}}
\newcommand{\xrtransformer}{\mytextsf{XR-Transformer}}
\newcommand{\lightxml}{\mytextsf{LightXML}}

\newcommand{\dexml}{\mytextsf{DEXML}}

\newcommand{\x}{\mathbf{x}}
\newcommand{\y}{\mathbf{y}}
\newcommand{\z}{\mathbf{z}}
\newcommand{\loss}{\mathcal{L}}
\newcommand{\w}{\mathbf{w}}
\newcommand{\ghat}{\widehat{\mathbf{g}}}
\newcommand{\g}{\mathbf{g}}
\newcommand{\W}{\mathbf{W}}
\newcommand{\f}{f}
\newcommand{\embsize}{d}
\newcommand{\encparam}{\boldsymbol{\theta}}
\newcommand{\enc}{\mathcal{E}_{\encparam}}
\newcommand{\enct}{\mathcal{E}_{\encparam^{(t)}}}
\newcommand{\enctp}{\mathcal{E}_{\encparam^{(t')}}}
\newcommand{\hardneg}{\mathcal{H}}
\newcommand{\randneg}{\mathcal{R}}
\newcommand{\totalneg}{\mathcal{N}}
\newcommand{\pos}{\mathcal{P}}
\newcommand{\randnegsize}{k_r}
\newcommand{\hardnegsize}{k_h}
\newcommand{\positivesize}{k_p}
\newcommand{\sota}{state-of-the-art}
\newcommand{\taus}{\tau_{s}}
\newcommand{\taur}{\tau_{r}}
\newcommand{\lr}{$lr$}
\newcommand{\batchsize}{S}

\usepackage{subfig}
\usepackage{algorithmic}
\usepackage{booktabs}
\usepackage{multirow}
\usepackage{graphicx}
\usepackage{amsmath}
\usepackage{mathtools}
\usepackage{amsthm}
	
\usepackage{color, colortbl}
\usepackage[linesnumbered,ruled,vlined]{algorithm2e}

\definecolor{Gray}{gray}{0.9}
\theoremstyle{plain}
\newtheorem{theorem}{Theorem}[section]

\newtheorem{lemma}[theorem]{Lemma}

\theoremstyle{definition}

\newtheorem{assumption}[theorem]{Assumption}
\theoremstyle{remark}

\usepackage[textsize=tiny]{todonotes}

\begin{document}
\title{\lexus{}: Accurate and Scalable ANNS-based Training of Extreme Classifiers}


\author{%
  Sonu Mehta  \\
Microsoft Research and IIT Delhi\\
  India \\
  \texttt{someh@microsoft.com} \\
  \And
 Jayashree Mohan  \\
Microsoft Research \\
  India \\
  \texttt{jamohan@microsoft.com} \\
  \AND
  Nagarajan Natarajan  \\
Microsoft Research \\
  India \\
  \texttt{nagarajn@microsoft.com} \\
  \And
  Ramachandran Ramjee \\
Microsoft Research \\
  India \\
  \texttt{ramjee@microsoft.com} \\
  \And
 Manik Varma \\
Microsoft Research \\
  India \\
  \texttt{manik@microsoft.com} \\
}

\maketitle

\begin{abstract}
  ``Extreme Classification'' (or XC) is the task of annotating data points (queries) with relevant labels (documents), from an extremely large set of $L$ possible labels, arising in search and recommendations. The most successful deep learning paradigm that has emerged over the last decade or so for XC is to embed the queries (and labels) using a deep encoder (e.g. DistilBERT), and use linear classifiers on top of the query embeddings. This architecture is of appeal because it enables millisecond-time inference using approximate nearest neighbor search (ANNS). The key question is how do we design training algorithms that are accurate as well as scale to $O(100M)$ labels on a limited number of GPUs.

State-of-the-art XC techniques that demonstrate high accuracies (e.g., \textcolor{black}{DEXML}, Ren\'ee, DEXA) on standard datasets have per-epoch training time that scales as $O(L)$ or employ expensive negative sampling strategies, which are prohibitive in XC scenarios. In this work, we develop an accurate and scalable XC algorithm ASTRA with two key observations: (a) building ANNS index on the classifier vectors and retrieving hard negatives using the classifiers aligns the negative sampling strategy to the loss function optimized; (b) keeping the ANNS indices current as the classifiers change through the epochs is prohibitively expensive while using stale negatives (refreshed periodically) results in poor accuracy; to remedy this, we propose a negative sampling strategy that uses a mixture of importance sampling and uniform sampling. By extensive evaluation on standard XC as well as proprietary datasets with 120M labels, we demonstrate that ASTRA achieves SOTA precision, while reducing training time by 4x-15x relative to the second best.
\end{abstract}






\section{Introduction}
\label{sec:intro}

We consider the \textit{extreme classification} (XC) problem of learning to retrieve relevant labels or documents, from an extremely large corpus, for a given search phrase or query. The problem has been an active sub-field of ML research for decades now as (a) several search and recommendation problems can be formulated as a supervised learning problem with an extremely large label space; 
and (b) it has an enormous revenue impact on businesses, e.g., identify ads to be shown from the hundred million keywords that are bid on, when a user issues a query on the search engine.

One of the most successful paradigms for XC \cite{Dahiya21,Dahiya23,Dahiya21b,dexapaper,DeepXMLURL,renee_2023} employs a deep encoder architecture for embedding the (text) queries, and, possibly also, labels; and then a linear one-vs-all style classifier layer is applied to the embeddings to produce the final predictions for the query, i.e., the score for a (query $\x$, label $\ell$) pair is the dot product of the embedding of $\x$ and the classifier vector $\w_\ell$. This paradigm is appealing because inference can be done in a few milliseconds on a CPU even with hundreds of millions of labels --- by leveraging approximate nearest neighbor search (ANNS) on the trained classifier vectors to retrieve the top-$k$ relevant labels for a given query~\cite{diskann-github}. \textcolor{black}{Given a dataset with $N$ training points and $L$ labels,} these methods crucially rely on meticulous negative mining strategies (such as clustering all the query embeddings periodically) in order to keep the per-epoch costs to $O(N*\log L)$ ($\log L$ is the number of mined negative labels per query) rather than $O(N*L)$ ($L$ is the default number of negative labels per query). This impacts the training time significantly when the number of labels scales to tens of millions (Section \ref{sec:results}), \textcolor{black}{especially since $N$ is typically of the same order or even larger than $L$}.

On the other hand, recent XC algorithms like \renee{} \cite{renee_2023} have successfully demonstrated how to \textit{jointly train} encoder parameters and classifiers to achieve state-of-the-art accuracies on standard XC datasets. However, the training time of \renee{} scales as $O(N*L)$ as it eschews negative sampling altogether. Therefore, scaling to tens of millions of labels with \renee{} is also expensive (Section \ref{sec:results}). 

\textcolor{black}{Recently, dual-encoder models like \dexml{} \cite{gupta2024dualencoders} have achieved state-of-the-art results on some XC datasets by using a decoupled softmax loss function. But, the lift in accuracy for \dexml{} comes at the cost of $O(N*L)$ training time, which prevents it from scaling to datasets beyond a few million labels.}


Thus, an open question is \textit{can we train extreme classifiers keeping the per-epoch time to $O(N*\log L)$ rather than $O(N*L)$, that would enable us scale to scenarios with hundreds of millions of labels, without compromising on the accuracy?} We propose \lexus, a novel approach for jointly learning the deep encoder parameters and the extreme classifiers, formulated based on two main observations.

First, the negative mining strategy should be aligned with the training loss function; to achieve this, we have to index on the \textit{classifier weights} (instead of label embeddings~\cite{xiong2020approximate,rocketqa} which is standard in literature) and \textit{retrieve hard negatives using the indexed classifier weights}. However, this is extremely challenging to realize in practice because of the overhead of keeping the ANNS indices up-to-date with the parameter changes during training.  Second, using stale indices (i.e., update indices every few epochs, and use stale hard negatives through the interim epochs) leads to poor accuracy as we show in Section \ref{sec:negsampling}. The theory of importance sampling \cite{johnson2018training} dictates that hard negatives is optimal for reducing the variance in the gradient estimates for SGD updates but empirical evidence suggests that it is no longer the case when the gradient estimates are computed using stale indices. This motivates us to design a simple, effective, \textit{and} fast negative mining strategy --- \lexus{} uses a mixture of stale importance sampling (hard negatives) and uniform sampling (random negatives) at each iteration. We show that this mixed strategy is both efficient and achieves high accuracy. We further show that the \lexus{} algorithm converges to a first-order stationary point at the rate of $O(1/T)$ after $T$ iterations. 

\textcolor{black}{Combining random with hard negatives has been used previously to either heuristically improve performance~\cite{Dahiya21} or to mitigate unstable gradients~\cite{10.1007/s10994-023-06468-w}. The novelty of ASTRA is in the insights for why a combination of random and hard negatives is needed (Section \ref{sec:negsampling}), a thorough analysis of its performance benefits, and strong empirical results on large-scale XC datasets.}


To the best of our knowledge, we are the first to demonstrate that ANNS-based training of extreme classifiers is effective at the scale of hundreds of millions of labels. For instance, on a proprietary dataset with 120M labels and 370M queries, \lexus{} achieves Precision@1 of 83.4 in 25 hours on 8 V100s. \renee, a state-of-the-art XC algorithm, achieves 83.8 Precision@1 but takes 375 hours or $15\times$ longer than \lexus{} to train on the same hardware. Implementations of other state-of-the-art XC techniques \cite{Dahiya23,dexapaper,gupta2023elias} simply do not scale to this size. We also evaluate \lexus{} on a number of publicly available datasets with up to 3M labels; it achieves comparable or better accuracy than state-of-the-art approaches like \renee{} or \dexml{} \textcolor{black}{while being $4.5\times$--$6.4\times$ faster compared to \renee{} and $14.6\times$--$80.4\times$ faster compared to \dexml{}.}

Our contributions are: \textbf{(1)} ANNS-based training algorithm for extreme classifiers that can scale to hundreds of millions of labels; \textbf{(2)} \lexus{} algorithm and proof that it converges to a first-order stationary point; \textbf{(3)} \lexus{} implementation that matches the state-of-the-art accuracy on datasets with up to 120 million labels, while reducing training times by up to $15\times$.

\section{eXtreme Classification: Setup and Challenges}
\label{sec:setup}
In the standard XC setup \citep{Huang13,Babbar17,Joulin17,Prabhu18b,Jain19,Khandagale19}, we are given a fixed set of $L$ labels, where $L$ can be hundreds of millions. We want to learn a prediction model $\f$ that outputs the \textit{most relevant} subset of labels for the input query $\x$. We have access to supervised data $\{ \x_i, \y_i \}_{i=1}^N$, where $\{ \ell: y_{i,\ell}=1\}$ are the subset of positive (or relevant) labels for the (text) query $\x_i$. 
We may also have access to textual description \citep{Mittal21,Dahiya21b,gupta2024dualencoders} or other meta-data (e.g., graphs \citep{Saini21,Mittal21b}) of the labels $\z_\ell$ in some scenarios.

\subsection{Deep XC model design and fast inference}
\label{sec:XCdesign}
The most successful paradigm of XC that has emerged over the last decade or so is formulating $f_\ell (\x) = \langle \w_\ell, \enc(\x) \rangle$ where $\enc: \x \mapsto \mathbb{R}^d$ is a deep encoder model, such as DistilBERT \cite{Sanh2019DistilBERTAD}, and $\w_\ell \in \mathbb{R}^d$ is the classifier vector \cite{Dahiya23,renee_2023, Dahiya21, Mittal21, Dahiya21b} (or embedding \cite{rocketqa, gupta2024dualencoders, dexapaper}) for label $\ell$ and  $d$ is the embedding dimension. At inference time, test query $\x$ is passed through the encoder $\enc(\x)$, and the top-$k$ scoring labels are computed via $f$. In real-time applications such as search and recommendations, this design enables millisecond-time inference, as we can leverage ANNS techniques to get the top-$k$ in time that is proportional to $k$ (say, 10) than the extremely large $L$.

\subsection{Challenges in scaling XC training} 
\label{sec:ANCE}

\textbf{Dual-Encoder models}: A popular XC approach is to jointly learn the embeddings of the queries and the documents (labels) using a deep encoder $\enc$. In each iteration, this involves computing the embeddings of both queries in the batch and all the labels. The encoder is learned using a contrastive loss that encourages the embeddings of queries to be close to their relevant documents. The learnt label embeddings are used for inference, i.e., $\w_\ell = \enc(\z_\ell)$. 

Mining useful negative labels is critical to drive the costs down to \textcolor{black}{$O(N\cdot |\encparam| \cdot \log L)$.} The go-to strategy is choosing the ``hardest negatives'' \cite{xiong2020approximate,gupta2024dualencoders}, i.e., (query, label) pair closest in the embedding space but whose ground-truth label is negative. We could leverage ANNS techniques for fast retrieval of the hard negatives but then keeping the ANNS indices current, as the embeddings change through the training epochs, adds significant overhead even when the number of labels is modest. 
\\ \\
\textbf{Classifier with Encoder models}: \textcolor{black}{Alternative XC approach is to not only learn the encoder (as above), but also one-vs-all style classifier vectors $\w_\ell \in \mathbb{R}^\embsize$ for all the labels. This approach improves accuracy at the expense of significant increase in the number of parameters. Per-epoch time of jointly learning the encoder and the classifiers scales as $\Omega(N(Ld + |\encparam|))$ and the memory requirement scales as $\Omega(Ld + |\encparam|)$, where $d$ is the embedding dimension, since all the classifiers need to be in memory.} 

\renee{} \cite{renee_2023}, state-of-the-art in this family, manages this joint training by leveraging multiple optimizations to alleviate both memory and compute bottlenecks, and using a hybrid data and model parallel training pipeline.
However, the per-epoch time of \renee{} still scales as $O(L)$, which implies slow convergence on datasets with a few tens to hundreds of millions of labels (Section \ref{sec:results}). Techniques like CascadeXML \cite{Kharbanda22} and ELIAS \cite{gupta2023elias} address scalability by learning representations at multiple resolutions or learning clustered indices of labels respectively. However, their accuracies are much worse (Section \ref{sec:results}). 

Finally, recent \textit{modular} training methods such as \ngame{} \cite{Dahiya23} and \dexa{} \cite{dexapaper} learn the encoder first, and then learn classifiers in the second stage using the fixed query embeddings. \textcolor{black}{While this staged training helps scalability to some extent, there are compute bottlenecks such as a) computing label embeddings at every iteration, and b) an expensive clustering procedure, for negative sampling, on all the queries $N$ which can be even larger than $L$ (Section \ref{sec:results}). For \ngame, balanced $k$-means clustering takes $O(NTk)$ time where $T, k$ are the number of iterations and clusters. The clustering time is significantly higher than building and retrieving from ANNS indices which is typically $O(N \log L)$, as $k \gg \log L$ (Section \ref{sec:anns_overheads}).}

\begin{figure*}
\centering
\begin{tabular}{@{}c@{}c@{\hspace{-2mm}}c@{}}

\scalebox{0.35}{
{\includegraphics[width=0.9\linewidth]{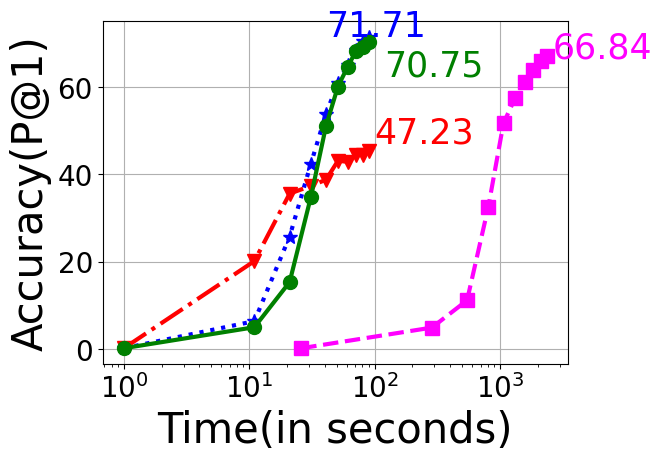} \label{fig:a}} 
}
&
\scalebox{0.5}{
{\includegraphics[width=0.9\linewidth]{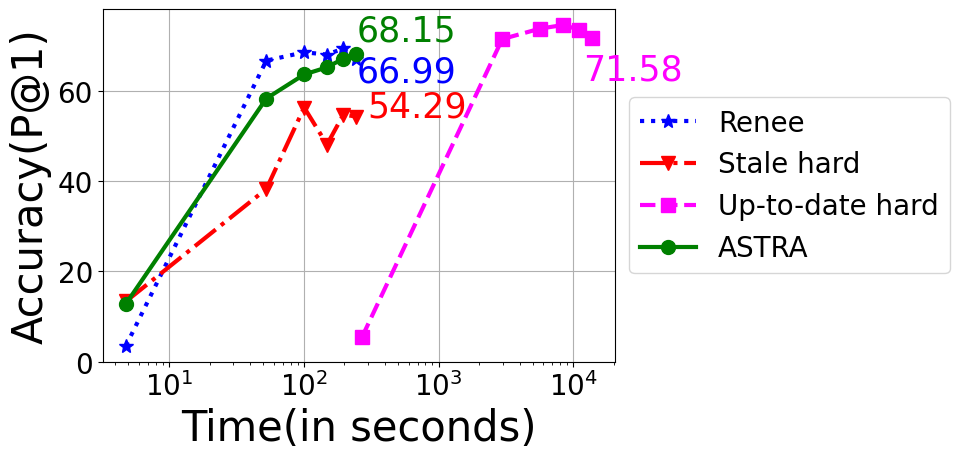} \label{fig:b}}
}
\\
(a) LF-AmazonTitles-1K & (b) LF-Amazon-1K
\end{tabular}
\caption[width=1\columnwidth]{Training time versus Precision@1 for different negative mining strategies: \textit{Stale hard} uses stale classifier weights, while \textit{Up-to-date hard} builds an ANNS index on classifier weights every iteration, to sample hard negatives. \lexus{} uses a mixture of random negatives and stale hard negatives (Section \ref{sec:negsampling}). The markers on the lines indicate epoch completion (in multiples of 5).  }
\label{fig:motivation}
\end{figure*}

\section{ANNS-based Training of eXtreme Classifiers}
\begin{figure*}
\resizebox{\textwidth}{!}{
\begin{tabular}{@{}c@{}c@{\hspace{-2mm}}c@{}}
\scalebox{0.75}{
\includegraphics[width=0.45\linewidth]{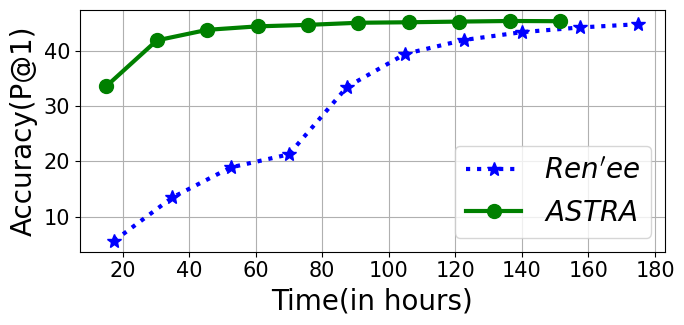}
}
&
\scalebox{0.75}{
\includegraphics[width=0.45\linewidth]{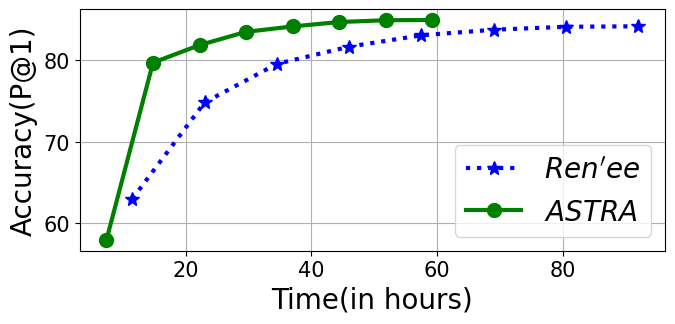}
}
&
\scalebox{0.75}{
\includegraphics[width=0.45\linewidth]{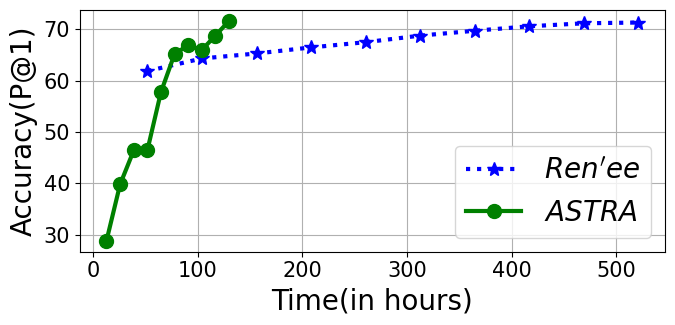}
\label{fig:epm20m}
}
\\
(a) LF-AmazonTitles-131K & (b) LF-Wikipedia-500K & (c) SS-20M

\end{tabular}
}
\vspace*{-5pt}
\caption{Training time versus P@1 for \renee{} and \lexus{} across 3 datasets. The markers on the lines indicate epoch completion (in multiples of 5).}
\label{fig:ttvsperf}
\end{figure*}

\label{sec:ASTRA}

We work in the XC setup introduced in Section \ref{sec:XCdesign}. Our model consists of two components: the encoder network $\enc : \x \mapsto \mathbb{R}^d$ and the one-vs-all classifiers $\w_\ell$, for $\ell \in [L]$; $[L]$ is short hand for the label set $\{1,2,\dots,L\}$. We seek to learn $f$ such that the top-$k$ retrieved labels using the scores $f_\ell(\x) = \w_\ell^T \enc(\x)$ for input query $\x$ are accurate.

\paragraph{Optimizing end-to-end XC loss:} 
\label{sec:endtoend}
Ideally, we want to train all the model parameters end-to-end, which is known to achieve state-of-the-art accuracies on XC datasets~\cite{renee_2023}. That is, we want to learn the classifiers $\w_\ell$ directly along with the encoder $\enc$ using a suitable loss on the predicted scores $f_\ell(\x_i)$ and the ground-truth labels $\y_i$. Standard loss functions used in XC training are contrastive loss such as the triplet loss \cite{Dahiya23,dexapaper} or the binary cross entropy (BCE) loss \cite{Dahiya21,zhai2023sigmoid,renee_2023}. In our work, we focus on the BCE loss given in equation \eqref{eqn:reneeloss} (Section~\ref{sec:theory}). The key challenge is to keep the per-epoch training time to $O(N \cdot \log L)$ rather than $O(N \cdot L)$, \textit{without} compromising on accuracy. 

\paragraph{Compute bottleneck:} Consider standard SGD updates with mini-batches to optimize the XC loss at each opoch. The per-epoch training time is dominated by backprop, i.e., computing the gradients of the loss w.r.t. encoder parameters and the $L$ classifier weights. For triplet loss, with all the $O(L)$ negative labels per query, per-epoch time will scale as $O(N(L \cdot \log L \cdot d  + |\encparam|))$, as the number of positive labels per query is typically $O(\log L)$. For BCE loss, it will scale as $O(N(Ld + |\encparam|))$, which is only slightly better. In either case, a major bottleneck is the dependence on $L$ (besides $N$), as that trumps all other factors in the extreme settings. The reason $L$ figures in the complexity is because the default number of negative labels per query is $O(L)$. The question then is if and how can we sample, say, $O(\log L)$ negative labels \textit{accurately} and \textit{efficiently} per query to remove the dependence on $L$.

\subsection{Proposed ANNS-based negative mining strategy}
\label{sec:negsampling}
Our negative mining strategy is motivated by two key observations and ideas.\\ 
\textbf{Observation 1: Align the negative mining strategy with the loss.} A naive solution is to directly apply ANCE-style hard negative sampling as proposed by \citet{xiong2020approximate} (and discussed in Section \ref{sec:ANCE}) for end-to-end training of the XC model. That is, embed the labels $\enc(\z_\ell)$, where $\z_\ell$ is the label meta-data, and create ANNS indices on these embeddings, and sample hard negatives for a given query from these embeddings. This is not ideal because the XC loss is computed on $\w_\ell^T\enc(\x_i)$, but the hardest negatives are retrieved using scores based on $\enc(\z_\ell)^T\enc(\x_i)$ --- this leads to the negative sampling strategy not being aligned with the loss. ANCE-style hard negative sampling techniques for training \textit{encoders} \cite{xiong2020approximate,Karpukhin20,rocketqa} are used in conjunction with the loss function defined on the scores $\enc(\z_\ell)^T\enc(\x_i)$. So, the right analogue for end-to-end XC training is to create ANNS \textit{indices on classifier weights} rather than label embeddings and \textit{retrieve hard negatives using the classifier weights} $\w_\ell$. This approach not only helps aligning the indices and the negative sampling strategy to the loss, but also has the side-benefit of enabling the solution to work on datasets without label features.\\ 
\textbf{Idea 1: Build ANNS-indices on classifier weights.} To the best of our knowledge, there is no work in the XC community that uses indices built on classifier weights for negative sampling and achieves high accuracy at the scale of millions of labels. State-of-the-art XC techniques like NGAME \cite{Dahiya23} and DEXA \cite{dexapaper} use the query embeddings to form mini-batches and then do in-batch negatives, while techniques like ANCE \cite{xiong2020approximate} and RocketQA \cite{rocketqa} create indices on the label embeddings. While it does appear natural to use classifier weights to do negative sampling, there is a significant hurdle to realize this in practice. This leads to our second key observation.\\ 
\textbf{Observation 2: Training with up-to-date negatives is prohibitively expensive while training with stale negatives results in poor accuracy.}
To understand how to ideally sample negatives, we can turn to the theory of importance sampling that guides optimal selection of mini-batches in SGD, i.e., how one should select mini-batches and learning rates to accelerate convergence of SGD. Proposition 3.1 of \citet{johnson2018training} suggests that we should select instances based on the norm of their gradients, in order to minimize the variance of the resulting gradients. We could apply the same theory to selecting negatives to estimate the loss function on a small set, instead of all negatives as in equation \eqref{eqn:reneeloss}, and accelerate the convergence of SGD. For a given query $\x$, the norm of the gradient of the loss w.r.t. $\w_\ell$ is proportional to the sigmoid of the score $f_\ell(\x) = \langle \w_{\ell},  \enc(\x)\rangle$.  Thus, the ``oracle'' importance sampling strategy for the query $\x$ at a given iteration $t$ is to sample label $\ell$ proportional to the sigmoid of the score $\langle\w^{(t)}_{\ell}, \enct(\x)\rangle$, ($^{(t)}$ denotes the latest iterate of the parameters). This requires re-creating ANNS-indices for $\w_{\ell}$ after every update to the parameters, which is very expensive. A practical implementation \cite{xiong2020approximate} is to use ``stale'' indices to do the sampling, i.e., at iteration $t$, we sample negative labels for query $\x$ using the importance sampling distribution that is offset by some (configurable number of) iterations. To be precise, this entails using the scores $\langle\w^{(t')}_{\ell}, \enctp(\x)\rangle$, where $t' < t$ denotes the last iteration when the indices were refreshed. Note that we could use fresh query embeddings, i.e, $\langle\w^{(t')}_{\ell}, \enct(\x)\rangle$, but in general, both can be stale, especially if we want to do this sampling asynchronously and not wait for the latest embeddings.

To understand the impact of staleness on performance, we consider small subsets of LF-AmazonTitles-131K and LF-Amazon-131K datasets by randomly sampling 1K labels and retaining all the queries that cover the 1K labels. We would like to compare the stale approach where we refresh ANNS indices every 5 epochs against the oracle sampling strategy where the ANNS indices are kept up-to-date by building fresh indices after every iteration. The oracle  strategy is computationally expensive even on the 1K sampled dataset and prohibitively expensive on the full 131K dataset. Note that for these datasets, the model size is dominated by the encoder network (over 60M parameters). We initialize the encoder with pre-trained weights (trained on the full train dataset), initialize classifiers randomly, and then jointly train the encoder and classifier parameters (with tuned learning rates) using mini-batched SGD updates and BCE loss. 

In Figure \ref{fig:motivation}, we show the convergence of SGD for the up-to-date and stale strategies. One epoch is a pass over all queries using mini-batches of size 512 and we train for 100 epochs in total. In both the cases, instead of sampling, we retrieve 200 top-scoring negative labels (i.e., hardest negatives) from the respective distributions per query. We see that, at convergence, the accuracy of the up-to-date strategy is very close to that of the state-of-the-art solution, \renee, that uses all the negatives as in equation \eqref{eqn:reneeloss}. But the up-to-date strategy is orders of magnitude slower that the other methods because of the sheer overhead in keeping the ANNS indices fresh with the parameter changes. On the other hand, we see that the stale strategy performs poorly. For example, on LF-AmazonTitles-1K, the converged stale solution (with a Precision@1 of 47.23) is significantly worse than the best (71.71). This observation also holds for LF-Amazon-1K as well as several XC datasets as we show in Section \ref{sec:ablation}, and motivates us to design a better sampling strategy.\\
\textbf{Idea 2: Use a mixture of importance sampling and uniform sampling distributions.}
Given a query $\x$, let $L_\x$ denote the positive label set of $\x$. We want to design a multinomial distribution over $[L] \setminus L_\x$ negative labels such that (a) it well approximates the aforementioned oracle sampling strategy, \textit{and importantly}, (b) it allows fast sampling. 

We draw inspiration from a relaxed implementation of the oracle importance sampling strategy considered by \citet{alain2015variance}; they propose to add a smoothing constant to the (stale) probabilities in order to be robust to variations induced by staleness in distributed SGD settings. In our experiments (Section \ref{sec:ablation}), we find that the na\"ive uniform sampling of negatives is often better than stale negative sampling strategy. So, to counter the impact of staleness, we propose a mixture distribution to sample negative labels for query $\x$ at iteration \textcolor{black}{ $t$: $P^{(t)} (\x) = (1-r)\ P^{(t')}_{\text{Imp}} (\x) + r\ P_{\text{Unif}}$, where $t' < t$ is the last ANNS index update iteration, $P_{\text{Imp}}$ and  $P_{\text{Unif}}$ are the importance and uniform negative sampling distributions respectively, and hyperparameter $r \in [0,1]$ governs the ratio of stale hard and uniformly random negatives. We show in Figure \ref{fig:motivation} that the proposed sampling strategy, labelled \lexus, indeed performs nearly as good as the best.
In \lexus{}, we implement this strategy efficiently as follows: for a given query, we retrieve $O(\log L)$ most-likely negatives based on $P^{(t')}_{\text{Imp}}$ and $c$ labels uniformly at random from $[L] \setminus L_\x$, where c is around 2000 for all datasets ( Section \ref{sec:results}).}

\begin{algorithm}[t]
  \caption{\lexus{}} 
  \label{algo:LEXUS}
  Init encoder $\enc(\x)$, classifiers $\W$, mini-batch size \batchsize, \# hard negatives $\hardnegsize$, \# random negatives $\randnegsize$, ANNS refresh interval $\tau_r$, epoch to start using hard negatives $\tau_s$, \# positives per query $k_p$  
  
  \For{epoch = 0,1, \dots}
  {
    Divide all data-points into random mini-batches of size $S$ \\
        \For{every mini-batch $S_t$}
        {
            Embed data-points (queries) using encoder  $\enct(\cdot)$. \\
            \If{$t <\tau_s$} {\{Use random negatives to train $\enc(\x)$ and  $\W$\} \\
            Sample $\totalneg_i$ negatives (=$\hardnegsize$+$\randnegsize$) uniformly at random from the feasible negative set for each data point $i \in S_t$}

            \If{$t >=\tau_s$ and $t \% \tau_r = 0$} {\{Redo ANNS refresh at regular intervals\} \\
                Use $\W^{(t)}$ to build an ANNS index on. \\
                Get $\hardnegsize$ nearest neighbours for each data-point using ANNS index, to be used until the next ANNS refresh}
            \If{$t >=\tau_s$}{\{Use hard+random negatives to train $\enc(\x)$ and  $\W$\} \\
            Get $\totalneg_i$ of negatives by sampling $\randnegsize$ negatives uniformly at random ($\randneg_i$) and $\hardnegsize$ hard negatives ($\hardneg_i$, most recent sample) for each data-point $i \in S_t$
            }
            
            Take positive labels $\pos_i$ (=$k_p$) and sampled negative labels $\totalneg_i$ for each data point $i \in S_t$  \\
            Compute BCE loss using $\pos_i$ and $\totalneg_i$ as given in equation \eqref{eqn:lexusloss} \\
            Update $\enct(\cdot)$ using mini-batch Adam over $S_t$ and  $\W^{(t)}$ using mini-batch SGD over $S_t$ 
            
        }
  
  }

\end{algorithm}

\subsection{ASTRA Algorithm}
\label{sec:algorithm}
We present the pseudocode of our proposed method, \lexus, in Algorithm \ref{algo:LEXUS}. For efficiency, as discussed above, we refresh the ANNS index once every $\tau_r$ epochs ($5$ in our experiments) and use the same set of negatives for the interim $\tau_r$ epochs. To further reduce the overheads, we do not wait until the refresh period completely lapses to retrieve the next set of negatives. That is, if we want to use the refreshed ANNS index in epoch 10, we start saving the query embeddings in epoch 8, build and retrieve nearest neighbours from ANNS index during epoch 9, so that by the time epoch 10 starts, we have the ``new'' set of negatives to be used for training. All of these operations can be done asynchronously on CPUs using an ANNS module while the training epochs are underway on GPUs. Thus, ANNS-based operations do not require any additional GPU compute or memory. We discuss the cost and accuracy of ANNS index building and retrieval in Section \ref{sec:anns_overheads}.

\textbf{Implementation:} \lexus{} implementation uses the Pytorch framework~\cite{paszke2019pytorch} and spans about 1200 lines of code. For smaller academic datasets, the implementation can be run on a single GPU, but when the number of labels are in a few millions or more, the memory requirements may be beyond the capabilities of a single GPU (e.g., 32GB V100). Therefore, following the implementation of \renee~\cite{renee_2023}, \lexus{} implementation has a hybrid data- and model-parallel architecture, where encoder is trained in a data-parallel fashion and the classifiers are trained in a model-parallel manner. 
\textcolor{black}{Similar to \renee{}, for small datasets with labels upto $3M$, we use embedding size $d = 768$; for very large datasets, we use a bottleneck layer that reduces $d$ to 64 and gradient accumulation to fit the required batch-sizes. We also augment the training data with label texts as training data-points in order to use label features for datasets wherever applicable. More details about these optimizations can be found in Appendix \ref{sec:implemenation_appendix}. An ablation of \lexus{} results with and without augmentation of training data with label texts is included in Table \ref{tab:data_aug}, Appendix \ref{sec:addresults}}.

\section{Convergence of \lexus}
\label{sec:theory}
Let $\W$ denote the $\embsize \times L$ matrix of classifier weights, $[L]$ denote the set of all labels, and let $\phi_\x = \enc (\x)$. For convenience, define $\loss_+(\encparam, \W; \x) := \sum_{\ell=1}^L y_{\ell} \log (1 + \exp(-\w_\ell^T\phi_\x))$, and $\loss_-(\encparam, \W; \x, \mathcal{S}) := \sum_{\ell \in \mathcal{S}} (1-y_{\ell})\big(\w_\ell^T\phi_\x + \log (1 + \exp(-\w_\ell^T\phi_\x))\big).$ The (full) loss function that we desire to optimize is the average of losses over data-points $\x$ given by: 
\begin{align}
\loss(\encparam, \W; \x) := \loss_+(\encparam, \W; \x) + \loss_-(\encparam, \W; \x, [L]) .
\label{eqn:reneeloss}
\end{align}
Consider \lexus{} at epoch $t$. Let $\hardneg^{(t)}_i$ denote the set of hard negatives for $\x_i$ sampled in Step 12, and $\randneg^{(t)}_i$ denote random negatives sampled in Step 15 of Algorithm \ref{algo:LEXUS}. The following lemma shows two properties: (i) the BCE loss estimator in Step 17, and given below in \eqref{eqn:lexusloss} for a given $\x$, is unbiased, as are its gradients; and (ii) the gradient estimator is strongly concentrated around its expectation. 
For ease, we will drop the superscripts from $\W^{(t)}$ and $\encparam^{(t)}$ when it is clear from the context.
\begin{align}
\loss^{(t)}_\lexus(\encparam, \W; \x) &:= \loss_+(\encparam, \W; \x)  + \loss_-(\encparam, \W; \x, \mathcal{H}^{(t)}) \nonumber\\ &+ \frac{1}{p \cdot \randnegsize} \loss_-(\encparam, \W; \x, \mathcal{R}^{(t)}),
\label{eqn:lexusloss}
\end{align}

where $p = 1/(L-\hardnegsize)$ is the probability of sampling a label uniformly at random from $[L] \setminus \mathcal{H}^{(t)}$.
\begin{lemma}
\begin{enumerate}
    \item Loss estimator (equation~\ref{eqn:lexusloss}) is unbiased, i.e., $$\mathbf{E}[\loss_\emph{\lexus}(\encparam, \W; \x)] = \loss(\encparam, \W; \x),$$ and so are the gradients computed in Step 18 of Algorithm \ref{algo:LEXUS}.
    \item For a given $\epsilon > 0$, $\delta > 0$, if $\randnegsize = O\big(\log(\frac{1}{\delta})\frac{1}{\epsilon^2}\big)$, letting $\| \cdot \|$ be the vectorized $L_2$ norm, with probability at least $1 - \delta$, 
$\|\nabla \loss_\emph{\lexus}(\encparam, \W; \x) - \nabla \loss(\encparam, \W; \x)\| \leq \epsilon \|\nabla \loss(\encparam, \W; \x) \|.$
\end{enumerate}
\label{lem:estimator}
\end{lemma}


With these properties in place, we can now give a formal convergence result for Algorithm \ref{algo:LEXUS}. In particular, the following theorem states that the algorithm converges to a first-order stationary point at the rate of $O(1/T)$, where the hidden (absolute) constants depend on the learning rate and the sub-optimality gap of the initial values for the model parameters. The proofs are in the Appendix \ref{sec:proofs}. 

\begin{table*}[]

\caption{Results (P$@k$ and Training Time in hours) on the proprietary datasets with 20M and 120M labels comparing \lexus{} with SOTA XC methods. The best results are in \textbf{bold}; the second best \underline{underlined}. Takeaways: \lexus{} matches the accuracy of SOTA \renee{} while being $4\times-15\times$ faster. NGAME is $2.3\times$ slower, 1\%-2\% less accurate than \lexus{} on SS-20M and does not scale to SS-120M due to the high cost of negative mining.\textcolor{red}{}}
\label{tab:rq1}

\centering
\resizebox{\textwidth}{!}{
\begin{tabular}{c|rrrr|crcc}
\hline
\multirow{2}{*}{Methods} & \multicolumn{4}{c|}{SS-20M \textbf{(N = 146M, V = 2 GPUs)}}                                                                            & \multicolumn{4}{c}{SS-120M \textbf{(N=370M, V= 8 GPUs)}}                                                                                    \\ \cline{2-9} 
                         & \multicolumn{1}{c}{P@1} & \multicolumn{1}{c|}{P@5}            & \multicolumn{1}{c|}{TT (hrs)} & \multicolumn{1}{c|}{Slowdown} & P@1                                & \multicolumn{1}{c|}{P@5}            & \multicolumn{1}{c|}{TT (hrs)} & Slowdown                \\ \hline
\ngame                    & 70.46                   & \multicolumn{1}{r|}{43.94}          & \multicolumn{1}{r|}{295.83}   & 2.3x                          & \multicolumn{2}{c|}{\textbf{--Not-scalable--}}                           & \multicolumn{1}{c|}{-}        & -                       \\
\renee                    & \underline{ 71.32}             & \multicolumn{1}{r|}{\textbf{46.64}} & \multicolumn{1}{r|}{520.83}   & 4x                            & \multicolumn{1}{r}{\textbf{83.78}} & \multicolumn{1}{r|}{\underline{ 41.27}}    & \multicolumn{1}{r|}{375.16}   & \multicolumn{1}{r}{15x} \\
\lexus                    & \textbf{71.62}          & \multicolumn{1}{r|}{\underline{ 46.60}}    & \multicolumn{1}{r|}{130.23}   & 1x                            & \multicolumn{1}{r}{\underline{ 83.37}}    & \multicolumn{1}{r|}{\textbf{41.67}} & \multicolumn{1}{r|}{25.04}    & \multicolumn{1}{r}{1x}  \\ \hline
\end{tabular}
}

\end{table*}

\begin{theorem}
Under certain smoothness assumptions (given in Appendix \ref{sec:assumption}) on the loss $\loss$ in \eqref{eqn:reneeloss}, and under the conditions of $\randnegsize, \epsilon$, and $\delta$ in Lemma \ref{lem:estimator}, for a given $\epsilon$, if the learning rate  $\eta = O(\frac{1-\epsilon}{1+\epsilon^2})$ in Step 18, Algorithm 1, after $T$ iterations, converges to $\encparam$, $\W$ s.t. $\|\nabla \loss(\encparam, \W)\| \leq O(1/T)$.
\label{thm:convergence}
\end{theorem}

\section{Experiments}
\label{sec:results}

\begin{table*}[]

\caption{Results (P@$k$ and Training Time in hours) on public datasets with label features comparing \lexus{} with SOTA XC methods. Results for \lightxml, \renee{} are from \citep{renee_2023}, \dexa{} and \ngame{}  are from \citep{dexapaper}; and \cascadexml{} and \elias{} numbers are reported from \citep{buvanesh2024enhancing}. The best results are in \textbf{bold}; the second best \underline{underlined}.  Takeways: \lexus{} performs comparable to SOTA across datasets with up-to 6.4$\times$ speed-up compared to \dexa{} and up-to 4.46$\times$ speed-up compared to \renee. \textcolor{black}{Note that the precision values reported are of single model and not for ensembles.}}
\label{tab:rq2_table_1}
\centering
\resizebox{\textwidth}{!}{
\begin{tabular}{c|rrrr|rrrl|rrrr}
\hline
\multirow{2}{*}{\textbf{Methods}} & \multicolumn{4}{c|}{\textbf{LF-AmazonTitles-131K}}                                                                                                         & \multicolumn{4}{c|}{\textbf{LF-AmazonTitles-1.3M}}                                                                                    & \multicolumn{4}{c}{\textbf{LF-Wikipedia-500K}}                                                                                        \\ \cline{2-13} 
                                  & \multicolumn{1}{c}{\textbf{P@1}} & \multicolumn{1}{c|}{\textbf{P@5}}   & \multicolumn{1}{c|}{\textbf{TT (hours)}} & \multicolumn{1}{l|}{\textbf{Slowdown}} & \multicolumn{1}{c}{\textbf{P@1}} & \multicolumn{1}{c|}{\textbf{P@5}}   & \multicolumn{1}{c|}{\textbf{TT (hours)}} & \textbf{Slowdown} & \multicolumn{1}{c}{\textbf{P@1}} & \multicolumn{1}{c|}{\textbf{P@5}}   & \multicolumn{1}{c|}{\textbf{TT (hours)}} & \textbf{Slowdown} \\ \hline
\dexml                              & 42.52                           & \multicolumn{1}{r|}{20.64}          & \multicolumn{1}{r|}{83.33}               & 41.25x                                  & \textbf{58.40}                      & \multicolumn{1}{r|}{\textbf{45.46}}          & \multicolumn{1}{r|}{$\sim$2000}              & 80.45x             & \textbf{85.78}                            & \multicolumn{1}{r|}{50.53}    & \multicolumn{1}{r|}{$\sim$592}               & 14.67x             \\                                  
\dexa                              & 45.78                            & \multicolumn{1}{r|}{21.29}          & \multicolumn{1}{r|}{13.01}               & 6.44x                                  & {55.76}                      & \multicolumn{1}{r|}{42.95}          & \multicolumn{1}{r|}{103.13}              & 4.15x             & {84.92}                            & \multicolumn{1}{r|}{50.51}    & \multicolumn{1}{r|}{57.51}               & 1.42x             \\
\ngame                             & 44.95                            & \multicolumn{1}{r|}{21.20}          & \multicolumn{1}{r|}{12.59}               & 6.23x                                  & 54.69                            & \multicolumn{1}{r|}{42.80}          & \multicolumn{1}{r|}{97.75}               & 3.93x             & 84.01                            & \multicolumn{1}{r|}{49.97}          & \multicolumn{1}{r|}{54.88}               & 1.36x             \\
\lightxml                          & 35.60                            & \multicolumn{1}{r|}{17.45}          & \multicolumn{1}{r|}{71.40}               & 35.35x                                 & \multicolumn{4}{c|}{--Not-scalable--}                                                                                                 & 81.59                            & \multicolumn{1}{r|}{47.64}          & \multicolumn{1}{r|}{185.56}              & 4.60x             \\
\elias                             & 37.28                            & \multicolumn{1}{r|}{18.14}          & \multicolumn{1}{r|}{4.33}                & 2.14x                                  & 47.48                            & \multicolumn{1}{r|}{38.6}           & \multicolumn{1}{r|}{40.00}               & 1.61x             & 81.94                            & \multicolumn{1}{r|}{48.75}          & \multicolumn{1}{r|}{138.67}              & 3.44x             \\
\multicolumn{1}{l|}{\cascadexml}   & 35.96                            & \multicolumn{1}{r|}{18.15}          & \multicolumn{1}{r|}{3.63}                & 1.80x                                  & 47.14                            & \multicolumn{1}{r|}{37.73}          & \multicolumn{1}{r|}{70.00}               & 2.82x             & 77.00                            & \multicolumn{1}{r|}{45.10}          & \multicolumn{1}{r|}{29.58}               & 0.73x             \\
\renee                             & \underline{46.05}                      & \multicolumn{1}{r|}{\textbf{22.04}} & \multicolumn{1}{r|}{2.05}                & 1.01x                                  & \underline{56.04}                   & \multicolumn{1}{r|}{{44.09}}    & \multicolumn{1}{r|}{27.33}               & 1.10x             & \underline{84.95}                   & \multicolumn{1}{r|}{\textbf{51.68}} & \multicolumn{1}{r|}{180.00}              & 4.46x             \\
\lexus                             & {\textbf{46.20}}             & \multicolumn{1}{r|}{\underline{21.95}}    & \multicolumn{1}{r|}{2.02}                & 1.00x                                  & {55.71}                      & \multicolumn{1}{r|}{\underline{44.26}} & \multicolumn{1}{r|}{24.86}               & 1.00x             & 84.88                            & \multicolumn{1}{r|}{\underline{51.30}}          & \multicolumn{1}{r|}{40.36}               & 1.00x             \\ \hline
\end{tabular}
}
\vskip -0.1in
\end{table*}

\textbf{Datasets:} We evaluate \lexus{} on multiple short-text and long-text datasets with and without label features from the Extreme Classification Repository \cite{XMLRepo}.  We also report results on proprietary datasets with 20M and 120M labels. These datasets cover a variety of applications including product-to-product recommendation (Amazon-670K, Amazon-3M, LF-Amazon-131K, LF-AmazonTitles-131K, and LF-AmazonTitles-1.3M), predicting Wikipedia categories (LF-Wikipedia-500K) and matching user queries to advertiser bid phrases in sponsored search (SS-20M, SS-120M). Please refer to Appendix \ref{sec:datasets_app} for more details on the datasets.
\\
\textbf{Baselines:} \textcolor{black}{For public datasets with label features, we compare with SOTA modular XC methods \dexa{} \citep{dexapaper},  \ngame{} \citep{Dahiya23}, end-to-end XC methods \lightxml{} \citep{Jiang21}, \elias{} \citep{Gupta22}, \cascadexml{} \citep{Kharbanda22}, \renee{} \citep{renee_2023} and dual-encoder based method \dexml{} \citep{gupta2024dualencoders}.}
More comprehensive baselines are covered in Appendix \ref{sec:addresults}.
Prior work employs ensembling to improve accuracy, which can be applied to \lexus{} as well. In this paper, we report numbers based on a single model. For proprietary datasets, we compare against \ngame{} and \renee; available implementations of other XC aforementioned  methods do not scale to $O(100M)$ labels.\\
\textbf{Hyperparameters:} We train \lexus{} using Adam and SGD optimizers for the encoder and classifiers respectively. We use cosine decay with warmup learning rate schedule. On the validation set, we tune \lexus's hyperparameters:  (i) batch-size, learning rate (\lr) (ii) for the encoder, (iii) for the classifiers, (iv) dropout, (v) weight decay for classifier, (vi) number of random negatives ($\randnegsize$), (vii) number of hard negatives ($\hardnegsize$), (viii) starting epoch for hard negative sampling ($\taus$), (ix) refresh frequency for ANNS index ($\taur$). \textcolor{black}{\lexus{} is fairly robust to the choice of hyperparameters (specific to our algorithm) $\taus$, $\taur$, $\randnegsize$ and $\hardnegsize$ as shown in Tables \ref{tab:ratio_ablation}, \ref{tab:tau_s_ablation} and \ref{tab:tau_r_ablation} (discussed in Section \ref{sec:anns_overheads}). Appendix \ref{sec:hyperparamtuning} has details on tuning and the chosen hyperparameter values.} 
The hyperparameters for baseline methods were set as per the respective papers. \\
 \textbf{Evaluation Metrics:} Models are evaluated using Precision@$k$ (P@$k$, $k \in {1, 5}$) and training time (All models are run on V100 GPUs).
 Results on other metrics, such as P@$3$, propensity-scored \citep{Jain16, Qaraei21, Schultheis22} variants of Precision@$k$ (PSP@$k$, $k \in {1, 3, 5}$), nDCG$@k$ (N@k) and PSN$@k$ \citep{Jain16, Qaraei21, Schultheis22} are reported in Appendix~\ref{sec:addresults}. Please refer to~\citep{XMLRepo} for definitions of all these metrics. 


\subsection{Accuracy and Efficiency of \lexus{} on the largest datasets}
\label{sec:rq1}

We evaluate the performance of \lexus{} on the two largest (proprietary) datasets with 20M and 120M labels. As mentioned early in this section, the available XC implementations that scale to these sizes are \renee{} and \ngame{}. Table~\ref{tab:rq1} shows that \textbf{\lexus{}} matches the \textit{\textbf{\sota{} (Precision@$k$, $k \in {1, 5}$) with 4$\times$ and 15$\times$ speed-up}} in the training time for SS-20M (where all the methods are run on 2 GPUs) and SS-120M (where all the methods are run on 8 GPUs) datasets respectively.

\begin{table}[]
\caption{Break-up of iteration time (in milliseconds) of \lexus{} and \renee{} for three datasets. *The optimized version of \renee{} does not calculate loss, hence marked as N/A.} 
 \label{tab:timebreakup}
\centering

\begin{tabular}{c|cc|cc|cc}
\hline
\textbf{Dataset}       & \multicolumn{2}{c|}{\textbf{LFAT-131K}} & \multicolumn{2}{c|}{\textbf{LFAT-1.3M}} & \multicolumn{2}{c}{\textbf{SS-20M}} \\ \hline
\textbf{Method}        & \textbf{\renee}     & \textbf{\lexus}     & \textbf{\renee}     & \textbf{\lexus}     & \textbf{\renee}   & \textbf{\lexus}   \\ \hline
\textbf{1.DataPrep}      & 0.1                & 0.2                & 0.1                & 0.2                & 0.1              & 0.2              \\
\textbf{2.CopyToGPU} & 0.3                & 0.3                & 0.3                & 0.3                & 0.3              & 0.3              \\
\textbf{3.EncFwdPass}    & 14.3               & 14.2               & 14.4               & 14.4               & 14.3             & 14.5             \\
\textbf{4.ClfFwdPass}    & \textit{0.6}       & \textbf{1.3}       & \textit{5.4}       & \textbf{1.4}       & \textit{76.8}    & \textbf{1.6}     \\
\textbf{5.LossCalc}      & N/A*               & 0.3                & N/A                & 0.3                & N/A              & 0.3              \\
\textbf{6.BwdPass}       & 20.4               & 21.9               & 39.6               & 31.6               & 183.3            & 32.2             \\ \hline
\textbf{7.Sum}           & 35.4               & 38.2               & 98.3               & 48.1               & 274.8            & 49.1  \\ \hline          
\end{tabular}

\end{table}

\subsection{Accuracy, Efficiency, and Memory Footprint Analysis of \lexus{} on public datasets}
\textbf{(a) Accuracy:} Table ~\ref{tab:rq2_table_1} compares \lexus{} with baseline and SOTA XC methods on three public datasets with label features. We observe that \lexus{} matches the SOTA Precision@$k$, $k \in \{1, 5\}$ on all the datasets \textcolor{black}{except P@$1$ for LF-AmazonTitles-1.3M}. Further, \lexus{} achieves up to 80$\times$ speed-up in training time compared to \dexml{} and up to 4.5$\times$ speed-up compared to \renee{} on 1 V100 GPU. Note that the precision values reported are of single model and not for ensembles \textcolor{black} {(Appendix \ref{sec:addresults} has ensemble results)}. We observe similar trends on datasets without label features presented in Appendix \ref{sec:addresults} due to space constraints.\\

\textcolor{black}{\textbf{(b) Efficiency:} As mentioned in Section~\ref{sec:negsampling}, the key advantage in training time comes from negative mining which reduces the per-epoch complexity to $O(N(\log L \cdot d + |\encparam|))$ from $O(N(Ld + |\encparam|))$ of \renee{}. The per-epoch computation of \lexus{} is dominated by the 6 steps shown in Table \ref{tab:timebreakup}. First, note that $O(\log L)$ vs $O(L)$ scaling is applicable only to the classifier forward (and the overall backward) pass steps, i.e., rows 4 and 6, and the loss calculation (row 5). Clearly, classifier forward and backward pass are the dominating values for large values of $L$ as seen from Table \ref{tab:timebreakup}; and we see that \renee’s compute scales linearly in $L$, as we go from left to right in the Table. On the other hand, for \lexus{}, the classifier forward pass remains more or less the same (around 1.5 ms), with very small variation across $L$. $O(\log L)$ scaling is not apparent here because the compute in steps 4 and 6 is dominated not by the $\log L$ sampled hard negatives per query, but by the constant but relatively larger number of random negatives per query. Because we use a constant number of random negatives per query (around 2000, $>> O(\log L)$ and independent of $L$, as shown in Table \ref{tab:hyperparam} of Appendix \ref{sec:hyperparamtuning}), the compute remains almost constant for \lexus{} in Table \ref{tab:timebreakup}, from left to right. This represents the key efficiency gain of \lexus{} compared to \renee, while also being competitive or even outperforming \renee{} in the extreme regime, as seen from Tables \ref{tab:rq1} and \ref{tab:rq2_table_1}.}

\textcolor{black}{Practical applications have a training budget. Figure \ref{fig:ttvsperf} shows that \lexus{} converges much faster compared to \renee{} consistently across datasets. }
For smaller $L$ values, the overall speed-up is not significant for aforementioned reasons.\\
\textbf{(c) Memory Footprint:} Table \ref{tab:memory} compares CPU and GPU memory usage of \lexus{} with \renee . Overall, negative sampling in \lexus{} leads to upto 668$\times$ reduction in GPU memory compared to \renee. For the classifier forward pass, \renee{} uses $O(\batchsize  \cdot L)$ GPU memory to save the output of matrix multiplications where \batchsize{} is mini batch-size. \lexus{} uses $O(\batchsize \cdot \log L)$ memory as the matrix multiplication is done with positives and selected negatives.

\textcolor{black}{
\subsubsection{ANNS - Overheads, Accuracy and Efficiency}
\label{sec:anns_overheads}
ANNS index is built on CPU and the retrieved negatives are stored as memmap files, after which the ANNS index is discarded. Memmap files are loaded into RAM during training. Table \ref{tab:memory} (`CPU' column) shows that the additional CPU memory usage because of these memmap files is very low. As described in Section \ref{sec:algorithm}, \lexus{} does not use any additional GPU memory to save or retrieve from ANNS indices.}\\ \\
\textcolor{black}{a) \textbf{Accuracy of ANNS}: To investigate the accuracy of sampled hard negatives, we compare two ANNS indices, DiskANN \cite{Subramanya19} and Faiss \cite{douze2024faiss}, in \lexus. The recall values of both these techniques are similar ($92.5\%-95\%$ across datasets), and in particular recall $> 90\%$ for both the techniques across datasets (recall is defined as the fraction of the true nearest neighbors retrieved by an ANNS algorithm). To further investigate the impact of ANNS choice in ASTRA, we trained \lexus{} using \textit{exact search} (recall is $100\%$, by definition) for small datasets (LF-AmazonTitles-131K, LF-Amazon-131K) instead of ANNS, keeping all other hyperparameters the same. Even in this exact setting, we do not observe any significant gains in terms of our final evaluation metrics on these datasets.}\\
\textcolor{black}{b) \textbf{Efficiency of ANNS}: Building ANNS index (e.g., DiskANN) typically costs $O(L \cdot \log L)$ time. The time complexity to retrieve $k$ nearest neighbours for $N$ queries from the ANNS index is $O(N \cdot k \cdot \log L)$. So, the total time to build and retrieve from ANNS index is $O(N \cdot \log L)$, since $N>L$ for most XC datasets (shown in Table 5 in Appendix \ref{sec:datasets_app}). For \lexus, the per-epoch training time scales as  $O(N(\log L \cdot d + |\encparam|))$; so, building and retrieving from ANNS index can be done within an epoch’s time, and we observe this in practice across datasets as well. 
}


\begin{table}[]
\caption{Memory usage (in MB) of \lexus{} and \renee{} on public datasets. Reduction in GPU memory usage is in parantheses.}  
 \label{tab:memory}
\centering

\begin{tabular}{c|rr|rr}
\hline
\multirow{2}{*}{\textbf{Dataset}} & \multicolumn{2}{c|}{\textbf{GPU}} & \multicolumn{2}{c}{\textbf{CPU}} \\ \cline{2-5} 
                              & \textbf{\renee} & \textbf{\lexus{}} & \textbf{\renee} & \textbf{\lexus{}} \\ \hline
\textbf{LF-Amazonitles-131K}            & 128.00         & 2.06 (62.2x)   & 0              & 1.35           \\
\textbf{LF-Amazon-131K}       & 128.00         & 1.91 (67.1x)   & 0              & 0.68           \\
\textbf{Amazon-670K}          & 327.19         & 3.57 (91.6x)   & 0              & 1.34           \\
\textbf{LF-AmazonTitles-1.3M} & 2549.35        & 3.81 (668.7x)  & 0              & 1.49           \\
\textbf{Amazon-3M}            & 1373.18        & 3.91 (351.5x)  & 0              & 1.62           \\ \hline
\end{tabular}
\end{table}

\begin{table*}[]
\centering
\caption{Ablative study (Section \ref{sec:negsampling}) of negative sampling strategies in \lexus{}: In all the rows, \# negative labels per query is fixed (to 2K). Hard negatives obtained using label embeddings are ``$\enc$-hard''; those obtained using classifiers are ``$\w$-hard''. 
\textit{All the hard negatives are stale (refreshed every 5 epochs)}. The best results are in \textbf{bold}; the second best \underline{underlined}.
}
\label{tab:rq3}
\vskip 0.1in
\resizebox{\textwidth}{!}{

\begin{tabular}{c|rr|rr|rr}
\hline
\multirow{2}{*}{\textbf{Methods}} & \multicolumn{2}{c|}{\textbf{LF-AmazonTitles-131K}} & \multicolumn{2}{c|}{\textbf{LF-Amazon-131K}} & \multicolumn{2}{c}{\textbf{LF-Wikipedia-500K}} \\ \cline{2-7} 
                                  & \textbf{P@1}          & \textbf{P@5}               & \textbf{P@1}          & \textbf{P@5}         & \textbf{P@1}           & \textbf{P@5}          \\ \hline
random                            & 44.57                 & \underline{ 21.83}                & 45.62                 & 22.9                 & 72.49                  & 44.58                 \\ \hline
$\enc$ hard                        & 41.98                 & 19.07                      & 15.39                 & 8.65                 & 57.37                  & 28.87                 \\
curriculum Learning ($\enc$ hard)               & \underline{45.84}           & 21.75                      & 43.81        & 21.25       & { 74.35}            & {46.24}        \\
random + $\enc$ hard               & 45.55                 & 21.55                      & 45.79                 & 22.15                & 81.40                  & 47.60                 \\ \hline
$w$ hard                          & 41.74                 & 20.79                      & 17.86                 & 8.87                 & 52.84                  & 22.92                 \\
curriculum Learning ($w$ hard)               & \underline{ 45.83}           & 21.58                      & \textbf{47.33}        & \textbf{22.47}       & \underline{ 84.56}            & \textbf{51.72}        \\
\rowcolor{Gray}
random + $w$ hard (\lexus)         & \textbf{46.20}        & { \textbf{21.95}}       & \underline{ 47.13}           & \underline{ 22.35}          & \textbf{84.88}         & \underline{ 51.30}           \\ \hline
\end{tabular}

}

\vskip -0.1in
\end{table*}

\subsection{Ablative study on the negative sampling strategy of \lexus{}}
\label{sec:ablation}
We perform ablative study on the two design choices of \lexus{} presented in Section \ref{sec:negsampling}. \textbf{(1)} \textit{building the ANNS index on the classifier weights}: A natural question is what if we built indices and sampled negatives using \textit{the label embeddings} instead, keeping all other choices in Algorithm \ref{algo:LEXUS} fixed. We observe from the Table~\ref{tab:rq3} that the proposed strategy of using classifier weights outperforms using label embeddings (in rows 2 and 3, denoted by $\enc$-hard) across three datasets. \textbf{(2)} \textit{the proposed mixture of stale hard and uniform distribution for sampling negatives}: We make two observations in this regard. First, we see that the proposed mixture distribution (last row) significantly outperforms using stale hard negatives (denoted by $w$-hard). This reinforces our observations in Section \ref{sec:negsampling} and Figure \ref{fig:motivation}. Second, the mixture distribution helps even when we use label embeddings for sampling, comparing rows 2 and 3. Finally, we compare curriculum learning, i.e., we start with random negatives and gradually increase the ratio of stale hard and random negatives through the epochs. We observe that the curriculum learning using classifier weights outperforms using label embeddings.

\section{Related Work}
\label{sec:related}
\begin{table}[]
\caption{Performance of \lexus{} with different values of $k_h, k_r$ (sampled hard and random negatives) on two datasets.} 
 \label{tab:ratio_ablation}
\centering

\begin{tabular}{r|rr|rr}
\hline
\multicolumn{1}{c|}{\multirow{2}{*}{\textbf{\begin{tabular}[c]{@{}c@{\textbf{$k_h,k_r$}}}\\ \end{tabular}}}} &
  \multicolumn{2}{c|}{\textbf{LF-AmazonTitles-131K}} &
  \multicolumn{2}{c}{\textbf{LF-Wikipedia-500K}} \\ \cline{2-5} 
\multicolumn{1}{c|}{} &
  \textbf{P@1} &
  \textbf{P@5} &
  \textbf{P@1} &
  \textbf{P@5} \\ \hline
50:400 & \textbf{46.20} & \textbf{21.95} & \textbf{84.88} & \textbf{51.30} \\
50:300 & 46.25 & 21.92 & 84.14 & 50.77 \\
50:200 & 46.23 & 21.89 & 84.10 & 50.21 \\
50:100 & 46.17 & 21.79 & 83.07 & 49.67 \\
30:400 & 46.15 & 21.91 & 84.09 & 50.70 \\
30:100 & 46.12 & 21.75 & 84.02 & 50.67 \\
20:400 & 46.10 & 21.92 & 84.03 & 50.63 \\
20:300 & 46.12 & 21.89 & 84.01 & 50.64 \\ \hline
\end{tabular}

\end{table}

We highlight XC techniques, especially negative mining, not addressed thus far. \\ 
\textbf{In-batch negatives:} In this strategy, for a given query in a mini-batch, the positive labels of other queries in the mini-batch are treated as its negatives. This is one of the most popular strategies for representation learning in the unsupervised~\citep{Faghri18,Lee19,Chen20,He20} as well as in the (partially) supervised learning settings \citep{Karpukhin20,khosla2020supervised}. \citet{Dahiya21b} use a variant of this approach, by forming mini-batches on \textit{labels} and mining the hardest few queries within the mini-batch instead. \\ 
\begin{table}[]
\caption{Performance of \lexus{} with varying $\tau_s$ (the starting epoch for hard negative sampling) on two datasets.} 
 \label{tab:tau_s_ablation}
\centering

\begin{tabular}{l|rr|rr}
\hline
\multicolumn{1}{c|}{\multirow{2}{*}{\textbf{$\tau_s$}}} & \multicolumn{2}{c|}{\textbf{LF-Amazon-131K}} & \multicolumn{2}{c}{\textbf{LF-AmazonTitles-1.3M}} \\ \cline{2-5} 
\multicolumn{1}{c|}{} & \textbf{P@1}   & \textbf{P@5}   & \textbf{P@1}   & \textbf{P@5}   \\ \hline
5                     & \textbf{47.13} & \textbf{22.35} & \textbf{55.71} & \textbf{44.26} \\
10                    & 46.29          & 21.96          & 52.17          & 41.40          \\
20                    & 46.24          & 21.97          & 51.76          & 41.28          \
\\ \hline
\end{tabular}

\end{table}

\begin{table}[]
\caption{Performance of \lexus{} with different values of $\tau_r$ (refresh frequency for ANNS index) on two datasets.} 
 \label{tab:tau_r_ablation}
\centering

\begin{tabular}{l|rr|rr}
\hline
\multirow{2}{*}{\textbf{$\tau_r$}} & \multicolumn{2}{c|}{\textbf{LF-Amazon-131K}} & \multicolumn{2}{l}{\textbf{LF-AmazonTitles-1.3M}} \\ \cline{2-5} 
                                 & \textbf{P@1}          & \textbf{P@5}         & \textbf{P@1}            & \textbf{P@5}            \\ \hline
5                                & \textbf{47.13}        & \textbf{22.35}       & \textbf{55.71}          & \textbf{44.26}          \\
10                               & 46.28                 & 21.95                & 50.43                   & 40.04                   \\
20                               & 46.02                 & 21.84                & 47.82                   & 29.06                   \\ \hline
\end{tabular}

\end{table}

\textbf{(Semi-)Global retrieval:} Here, the negatives are retrieved globally from the full label set \citep{Guo19,Reddi19,Karpukhin20,Kumar17,xiong2020approximate}. 
\citet{xiong2020approximate} propose ANCE method, that retrieves hard negatives via a global index of labels that is asynchronously updated (every few iterations). However, as we show in Section \ref{sec:ablation}, this strategy compromises on the accuracy. In the \textit{stochastic negative mining} strategy proposed by \citet{Reddi19}, a batch $B$ of negatives is first sampled uniformly at random globally, and then the top-$k$ hardest negatives are obtained from within the batch, where typically $k \ll B$. To strike a fine balance between memory- and compute-efficiency, and accuracy, semi-global retrieval strategies have been proposed \citep{rocketqa,hofstatter21,Dahiya23}. \citet{rocketqa} propose a ``cross-batch'' mining strategy for data-parallel training on multiple GPUs, where they include the negatives from the other mini-batches (on different GPUs). Further, to mitigate the risk of including false negatives, they train a cross-encoder to filter out potential false negatives, and retain only the high-scoring (hard) negatives. \\
\textbf{Streaming/caching strategies:} ~\citet{He20} maintain a streaming label cache (negatives) for updating the encoder parameters. ~\citet{lindgren2021efficient} use the Gumbel-Max technique to sample negatives from the softmax distribution. For efficiency, they first sample a large multi-set of items (``negative cache''), and update the oldest few embeddings in the cache at every iteration.

\section{Conclusions, Limitations and Future Work}
\label{sec:conclusion}
We show that it is indeed viable to do ANNS-based training of extreme classifiers in a scalable fashion, and achieve SOTA accuracies. The prevailing wisdom in the information retrieval community of using hard (but stale) negatives to train encoders does not extend to XC training as we demonstrate in this work. We motivate our approach using empirical observations, guided by theoretical insights from the optimization community. We show that \lexus{} scales to 100M labels with an order of magnitude less training time than the previous SOTA, while matching the accuracies. One limitation of our approach is that scaling to $L = O(1B)$ brings a new set of challenges because the ANNS indexing will start to become the bottleneck. Solving this challenge might require a new algorithm for ANNS or perhaps a completely new approach for XC.  We believe similar ANNS-based training strategies can be more broadly applied, for instance, to scaling NGAME \citep{Dahiya23} and DEXA \cite{dexapaper}, which opens up exciting new frontiers for further research in XC.

\newpage
\bibliographystyle{ACM-Reference-Format}
\bibliography{paper}


\begin{thebibliography}{47}


\ifx \showCODEN    \undefined \def \showCODEN     #1{\unskip}     \fi
\ifx \showDOI      \undefined \def \showDOI       #1{#1}\fi
\ifx \showISBNx    \undefined \def \showISBNx     #1{\unskip}     \fi
\ifx \showISBNxiii \undefined \def \showISBNxiii  #1{\unskip}     \fi
\ifx \showISSN     \undefined \def \showISSN      #1{\unskip}     \fi
\ifx \showLCCN     \undefined \def \showLCCN      #1{\unskip}     \fi
\ifx \shownote     \undefined \def \shownote      #1{#1}          \fi
\ifx \showarticletitle \undefined \def \showarticletitle #1{#1}   \fi
\ifx \showURL      \undefined \def \showURL       {\relax}        \fi
\providecommand\bibfield[2]{#2}
\providecommand\bibinfo[2]{#2}
\providecommand\natexlab[1]{#1}
\providecommand\showeprint[2][]{arXiv:#2}

\bibitem[Alain et~al\mbox{.}(2015)]%
        {alain2015variance}
\bibfield{author}{\bibinfo{person}{Guillaume Alain}, \bibinfo{person}{Alex Lamb}, \bibinfo{person}{Chinnadhurai Sankar}, \bibinfo{person}{Aaron Courville}, {and} \bibinfo{person}{Yoshua Bengio}.} \bibinfo{year}{2015}\natexlab{}.
\newblock \showarticletitle{Variance reduction in sgd by distributed importance sampling}.
\newblock \bibinfo{journal}{\emph{arXiv preprint arXiv:1511.06481}} (\bibinfo{year}{2015}).
\newblock


\bibitem[Anonymous(2019)]%
        {DeepXMLURL}
\bibfield{author}{\bibinfo{person}{Anonymous}.} \bibinfo{year}{2019}\natexlab{}.
\newblock \showarticletitle{{Source code and datasets for DeepXML}}. In \bibinfo{booktitle}{\emph{Anonymous}}.
\newblock


\bibitem[Babbar and Sch\"{o}lkopf(2017)]%
        {Babbar17}
\bibfield{author}{\bibinfo{person}{R. Babbar} {and} \bibinfo{person}{B. Sch\"{o}lkopf}.} \bibinfo{year}{2017}\natexlab{}.
\newblock \showarticletitle{{DiSMEC: Distributed Sparse Machines for Extreme Multi-label Classification}}. In \bibinfo{booktitle}{\emph{WSDM}}.
\newblock


\bibitem[Bhatia et~al\mbox{.}(2016)]%
        {XMLRepo}
\bibfield{author}{\bibinfo{person}{K. Bhatia}, \bibinfo{person}{K. Dahiya}, \bibinfo{person}{H. Jain}, \bibinfo{person}{P. Kar}, \bibinfo{person}{A. Mittal}, \bibinfo{person}{Y. Prabhu}, {and} \bibinfo{person}{M. Varma}.} \bibinfo{year}{2016}\natexlab{}.
\newblock \bibinfo{title}{{The Extreme Classification Repository: Multi-label Datasets \& Code}}.
\newblock
\newblock
\urldef\tempurl%
\url{http://manikvarma.org/downloads/XC/XMLRepository.html}
\showURL{%
\tempurl}


\bibitem[Buvanesh et~al\mbox{.}(2024)]%
        {buvanesh2024enhancing}
\bibfield{author}{\bibinfo{person}{Anirudh Buvanesh}, \bibinfo{person}{Rahul Chand}, \bibinfo{person}{Jatin Prakash}, \bibinfo{person}{Bhawna Paliwal}, \bibinfo{person}{Mudit Dhawan}, \bibinfo{person}{Neelabh Madan}, \bibinfo{person}{Deepesh Hada}, \bibinfo{person}{Vidit Jain}, \bibinfo{person}{SONU MEHTA}, \bibinfo{person}{Yashoteja Prabhu}, \bibinfo{person}{Manish Gupta}, \bibinfo{person}{Ramachandran Ramjee}, {and} \bibinfo{person}{Manik Varma}.} \bibinfo{year}{2024}\natexlab{}.
\newblock \showarticletitle{Enhancing Tail Performance in Extreme Classifiers by Label Variance Reduction}. In \bibinfo{booktitle}{\emph{The Twelfth International Conference on Learning Representations}}.
\newblock
\urldef\tempurl%
\url{https://openreview.net/forum?id=6ARlSgun7J}
\showURL{%
\tempurl}


\bibitem[Chen et~al\mbox{.}(2020)]%
        {Chen20}
\bibfield{author}{\bibinfo{person}{T. Chen}, \bibinfo{person}{S. Kornblith}, \bibinfo{person}{M. Norouzi}, {and} \bibinfo{person}{G. Hinton}.} \bibinfo{year}{2020}\natexlab{}.
\newblock \showarticletitle{{A simple framework for contrastive learning of visual representations}}. In \bibinfo{booktitle}{\emph{ICML}}.
\newblock


\bibitem[Dahiya et~al\mbox{.}(2021a)]%
        {Dahiya21b}
\bibfield{author}{\bibinfo{person}{K. Dahiya}, \bibinfo{person}{A. Agarwal}, \bibinfo{person}{D. Saini}, \bibinfo{person}{K. Gururaj}, \bibinfo{person}{J. Jiao}, \bibinfo{person}{A. Singh}, \bibinfo{person}{S. Agarwal}, \bibinfo{person}{P. Kar}, {and} \bibinfo{person}{M. Varma}.} \bibinfo{year}{2021}\natexlab{a}.
\newblock \showarticletitle{{SiameseXML: Siamese Networks meet Extreme Classifiers with 100M Labels}}. In \bibinfo{booktitle}{\emph{ICML}}.
\newblock


\bibitem[Dahiya et~al\mbox{.}(2023a)]%
        {Dahiya23}
\bibfield{author}{\bibinfo{person}{K. Dahiya}, \bibinfo{person}{N. Gupta}, \bibinfo{person}{D. Saini}, \bibinfo{person}{A. Soni}, \bibinfo{person}{Y. Wang}, \bibinfo{person}{K. Dave}, \bibinfo{person}{J. Jiao}, \bibinfo{person}{K. Gururaj}, \bibinfo{person}{P. Dey}, \bibinfo{person}{A. Singh}, \bibinfo{person}{D. Hada}, \bibinfo{person}{V. Jain}, \bibinfo{person}{B. Paliwal}, \bibinfo{person}{A. Mittal}, \bibinfo{person}{S. Mehta}, \bibinfo{person}{R. Ramjee}, \bibinfo{person}{S. Agarwal}, \bibinfo{person}{P. Kar}, {and} \bibinfo{person}{M. Varma}.} \bibinfo{year}{2023}\natexlab{a}.
\newblock \showarticletitle{NGAME: Negative mining-aware mini-batching for extreme classification}. In \bibinfo{booktitle}{\emph{WSDM}}.
\newblock


\bibitem[Dahiya et~al\mbox{.}(2021b)]%
        {Dahiya21}
\bibfield{author}{\bibinfo{person}{K. Dahiya}, \bibinfo{person}{D. Saini}, \bibinfo{person}{A. Mittal}, \bibinfo{person}{A. Shaw}, \bibinfo{person}{K. Dave}, \bibinfo{person}{A. Soni}, \bibinfo{person}{H. Jain}, \bibinfo{person}{S. Agarwal}, {and} \bibinfo{person}{M. Varma}.} \bibinfo{year}{2021}\natexlab{b}.
\newblock \showarticletitle{{DeepXML: A Deep Extreme Multi-Label Learning Framework Applied to Short Text Documents}}. In \bibinfo{booktitle}{\emph{WSDM}}.
\newblock


\bibitem[Dahiya et~al\mbox{.}(2023b)]%
        {dexapaper}
\bibfield{author}{\bibinfo{person}{Kunal Dahiya}, \bibinfo{person}{Sachin Yadav}, \bibinfo{person}{Sushant Sondhi}, \bibinfo{person}{Deepak Saini}, \bibinfo{person}{Sonu Mehta}, \bibinfo{person}{Jian Jiao}, \bibinfo{person}{Sumeet Agarwal}, \bibinfo{person}{Purushottam Kar}, {and} \bibinfo{person}{Manik Varma}.} \bibinfo{year}{2023}\natexlab{b}.
\newblock \showarticletitle{Deep Encoders with Auxiliary Parameters for Extreme Classification}. In \bibinfo{booktitle}{\emph{Proceedings of the 29th ACM SIGKDD Conference on Knowledge Discovery and Data Mining}} (<conf-loc>, <city>Long Beach</city>, <state>CA</state>, <country>USA</country>, </conf-loc>) \emph{(\bibinfo{series}{KDD '23})}. \bibinfo{address}{New York, NY, USA}, \bibinfo{pages}{358–367}.
\newblock


\bibitem[Douze et~al\mbox{.}(2024)]%
        {douze2024faiss}
\bibfield{author}{\bibinfo{person}{Matthijs Douze}, \bibinfo{person}{Alexandr Guzhva}, \bibinfo{person}{Chengqi Deng}, \bibinfo{person}{Jeff Johnson}, \bibinfo{person}{Gergely Szilvasy}, \bibinfo{person}{Pierre-Emmanuel Mazaré}, \bibinfo{person}{Maria Lomeli}, \bibinfo{person}{Lucas Hosseini}, {and} \bibinfo{person}{Hervé Jégou}.} \bibinfo{year}{2024}\natexlab{}.
\newblock \showarticletitle{The Faiss library}.
\newblock  (\bibinfo{year}{2024}).
\newblock
\showeprint[arxiv]{2401.08281}~[cs.LG]


\bibitem[Faghri et~al\mbox{.}(2018)]%
        {Faghri18}
\bibfield{author}{\bibinfo{person}{F. Faghri}, \bibinfo{person}{D.-J. Fleet}, \bibinfo{person}{J.-R. Kiros}, {and} \bibinfo{person}{S. Fidler}.} \bibinfo{year}{2018}\natexlab{}.
\newblock \showarticletitle{{VSE++: Improving Visual-Semantic Embeddings with Hard Negatives}}. In \bibinfo{booktitle}{\emph{BMVC}}.
\newblock


\bibitem[Guo et~al\mbox{.}(2019)]%
        {Guo19}
\bibfield{author}{\bibinfo{person}{C. Guo}, \bibinfo{person}{A. Mousavi}, \bibinfo{person}{X. Wu}, \bibinfo{person}{D.-N. Holtmann-Rice}, \bibinfo{person}{S. Kale}, \bibinfo{person}{S. Reddi}, {and} \bibinfo{person}{S. Kumar}.} \bibinfo{year}{2019}\natexlab{}.
\newblock \showarticletitle{{Breaking the Glass Ceiling for Embedding-Based Classifiers for Large Output Spaces}}. In \bibinfo{booktitle}{\emph{NeurIPS}}.
\newblock


\bibitem[Gupta et~al\mbox{.}(2022)]%
        {Gupta22}
\bibfield{author}{\bibinfo{person}{Nilesh Gupta}, \bibinfo{person}{Patrick~H. Chen}, \bibinfo{person}{Hsiang{-}Fu Yu}, \bibinfo{person}{Cho{-}Jui Hsieh}, {and} \bibinfo{person}{Inderjit~S. Dhillon}.} \bibinfo{year}{2022}\natexlab{}.
\newblock \showarticletitle{{ELIAS:} End-to-End Learning to Index and Search in Large Output Spaces}. In \bibinfo{booktitle}{\emph{Advances in Neural Information Processing Systems 35: Annual Conference on Neural Information Processing Systems 2022, NeurIPS 2022, New Orleans, LA, USA, November 28 - December 9, 2022}}, \bibfield{editor}{\bibinfo{person}{Sanmi Koyejo}, \bibinfo{person}{S.~Mohamed}, \bibinfo{person}{A.~Agarwal}, \bibinfo{person}{Danielle Belgrave}, \bibinfo{person}{K.~Cho}, {and} \bibinfo{person}{A.~Oh}} (Eds.).
\newblock
\urldef\tempurl%
\url{http://papers.nips.cc/paper\_files/paper/2022/hash/7d4f98f916494121aca3da02e36a4d18-Abstract-Conference.html}
\showURL{%
\tempurl}


\bibitem[Gupta et~al\mbox{.}(2024)]%
        {gupta2024dualencoders}
\bibfield{author}{\bibinfo{person}{Nilesh Gupta}, \bibinfo{person}{Devvrit}, \bibinfo{person}{Ankit~Singh Rawat}, \bibinfo{person}{Srinadh Bhojanapalli}, \bibinfo{person}{Prateek Jain}, {and} \bibinfo{person}{Inderjit~S. Dhillon}.} \bibinfo{year}{2024}\natexlab{}.
\newblock \showarticletitle{Dual-Encoders for Extreme Multi-label Classification}. In \bibinfo{booktitle}{\emph{The Twelfth International Conference on Learning Representations, {ICLR} 2024, Vienna, Austria, May 7-11, 2024}}. \bibinfo{publisher}{OpenReview.net}.
\newblock
\urldef\tempurl%
\url{https://openreview.net/forum?id=dNe1T0Ahby}
\showURL{%
\tempurl}


\bibitem[He et~al\mbox{.}(2020)]%
        {He20}
\bibfield{author}{\bibinfo{person}{K. He}, \bibinfo{person}{Haoqi Fan}, \bibinfo{person}{Yuxin W.}, \bibinfo{person}{S. Xie}, {and} \bibinfo{person}{R. Girshick}.} \bibinfo{year}{2020}\natexlab{}.
\newblock \showarticletitle{{Momentum contrast for unsupervised visual representation learning}}. In \bibinfo{booktitle}{\emph{CVPR}}.
\newblock


\bibitem[Hofstätter et~al\mbox{.}(2021)]%
        {hofstatter21}
\bibfield{author}{\bibinfo{person}{S. Hofstätter}, \bibinfo{person}{S.-C. Lin}, \bibinfo{person}{J.-H. Yang}, \bibinfo{person}{J. Lin}, {and} \bibinfo{person}{A. Hanbury}.} \bibinfo{year}{2021}\natexlab{}.
\newblock \showarticletitle{{Efficiently Teaching an Effective Dense Retriever with Balanced Topic Aware Sampling}}. In \bibinfo{booktitle}{\emph{SIGIR}}.
\newblock


\bibitem[Huang et~al\mbox{.}(2013)]%
        {Huang13}
\bibfield{author}{\bibinfo{person}{P.~S. Huang}, \bibinfo{person}{X. He}, \bibinfo{person}{J. Gao}, \bibinfo{person}{L. Deng}, \bibinfo{person}{A. Acero}, {and} \bibinfo{person}{L. Heck}.} \bibinfo{year}{2013}\natexlab{}.
\newblock \showarticletitle{{Learning Deep Structured Semantic Models for Web Search using Clickthrough Data}}. In \bibinfo{booktitle}{\emph{CIKM}}.
\newblock


\bibitem[Jain et~al\mbox{.}(2019)]%
        {Jain19}
\bibfield{author}{\bibinfo{person}{H. Jain}, \bibinfo{person}{V. Balasubramanian}, \bibinfo{person}{B. Chunduri}, {and} \bibinfo{person}{M. Varma}.} \bibinfo{year}{2019}\natexlab{}.
\newblock \showarticletitle{{Slice: Scalable Linear Extreme Classifiers trained on 100 Million Labels for Related Searches}}. In \bibinfo{booktitle}{\emph{WSDM}}.
\newblock


\bibitem[Jain et~al\mbox{.}(2016)]%
        {Jain16}
\bibfield{author}{\bibinfo{person}{H. Jain}, \bibinfo{person}{Y. Prabhu}, {and} \bibinfo{person}{M. Varma}.} \bibinfo{year}{2016}\natexlab{}.
\newblock \showarticletitle{{Extreme Multi-label Loss Functions for Recommendation, Tagging, Ranking and Other Missing Label Applications}}. In \bibinfo{booktitle}{\emph{KDD}}.
\newblock


\bibitem[Jain et~al\mbox{.}(2023)]%
        {renee_2023}
\bibfield{author}{\bibinfo{person}{Vidit Jain}, \bibinfo{person}{Jatin Prakash}, \bibinfo{person}{Deepak Saini}, \bibinfo{person}{Jian Jiao}, \bibinfo{person}{Ramachandran Ramjee}, {and} \bibinfo{person}{Manik Varma}.} \bibinfo{year}{2023}\natexlab{}.
\newblock \showarticletitle{Renee: End-to-end training of extreme classification models}.
\newblock \bibinfo{journal}{\emph{Proceedings of Machine Learning and Systems}} (\bibinfo{year}{2023}).
\newblock


\bibitem[Jiang et~al\mbox{.}(2021)]%
        {Jiang21}
\bibfield{author}{\bibinfo{person}{T. Jiang}, \bibinfo{person}{D. Wang}, \bibinfo{person}{L. Sun}, \bibinfo{person}{H. Yang}, \bibinfo{person}{Z. Zhao}, {and} \bibinfo{person}{F. Zhuang}.} \bibinfo{year}{2021}\natexlab{}.
\newblock \showarticletitle{{LightXML: Transformer with Dynamic Negative Sampling for High-Performance Extreme Multi-label Text Classification}}. In \bibinfo{booktitle}{\emph{AAAI}}.
\newblock


\bibitem[Johnson and Guestrin(2018)]%
        {johnson2018training}
\bibfield{author}{\bibinfo{person}{Tyler~B Johnson} {and} \bibinfo{person}{Carlos Guestrin}.} \bibinfo{year}{2018}\natexlab{}.
\newblock \showarticletitle{Training deep models faster with robust, approximate importance sampling}.
\newblock \bibinfo{journal}{\emph{Advances in Neural Information Processing Systems}}  \bibinfo{volume}{31} (\bibinfo{year}{2018}).
\newblock


\bibitem[Joulin et~al\mbox{.}(2017)]%
        {Joulin17}
\bibfield{author}{\bibinfo{person}{A. Joulin}, \bibinfo{person}{E. Grave}, \bibinfo{person}{P. Bojanowski}, {and} \bibinfo{person}{T. Mikolov}.} \bibinfo{year}{2017}\natexlab{}.
\newblock \showarticletitle{{Bag of Tricks for Efficient Text Classification}}. In \bibinfo{booktitle}{\emph{EACL}}.
\newblock


\bibitem[Karpukhin et~al\mbox{.}(2020)]%
        {Karpukhin20}
\bibfield{author}{\bibinfo{person}{V. Karpukhin}, \bibinfo{person}{B. Oguz}, \bibinfo{person}{S. Min}, \bibinfo{person}{P. Lewis}, \bibinfo{person}{L. Wu}, \bibinfo{person}{S. Edunov}, \bibinfo{person}{D. Chen}, {and} \bibinfo{person}{W.-T. Yih}.} \bibinfo{year}{2020}\natexlab{}.
\newblock \showarticletitle{Dense Passage Retrieval for Open-Domain Question Answering}. In \bibinfo{booktitle}{\emph{EMNLP}}.
\newblock


\bibitem[Khandagale et~al\mbox{.}(2020)]%
        {Khandagale19}
\bibfield{author}{\bibinfo{person}{S. Khandagale}, \bibinfo{person}{H. Xiao}, {and} \bibinfo{person}{R. Babbar}.} \bibinfo{year}{2020}\natexlab{}.
\newblock \showarticletitle{{Bonsai: diverse and shallow trees for extreme multi-label classification}}.
\newblock \bibinfo{journal}{\emph{ML}} (\bibinfo{year}{2020}).
\newblock


\bibitem[Kharbanda et~al\mbox{.}(2022)]%
        {Kharbanda22}
\bibfield{author}{\bibinfo{person}{S. Kharbanda}, \bibinfo{person}{A. Banerjee}, \bibinfo{person}{E. Schultheis}, {and} \bibinfo{person}{R. Babbar}.} \bibinfo{year}{2022}\natexlab{}.
\newblock \showarticletitle{CascadeXML: Rethinking Transformers for End-to-end Multi-resolution Training in Extreme Multi-label Classification}. In \bibinfo{booktitle}{\emph{NeurIPS}}.
\newblock


\bibitem[Khosla et~al\mbox{.}(2020)]%
        {khosla2020supervised}
\bibfield{author}{\bibinfo{person}{Prannay Khosla}, \bibinfo{person}{Piotr Teterwak}, \bibinfo{person}{Chen Wang}, \bibinfo{person}{Aaron Sarna}, \bibinfo{person}{Yonglong Tian}, \bibinfo{person}{Phillip Isola}, \bibinfo{person}{Aaron Maschinot}, \bibinfo{person}{Ce Liu}, {and} \bibinfo{person}{Dilip Krishnan}.} \bibinfo{year}{2020}\natexlab{}.
\newblock \showarticletitle{Supervised contrastive learning}.
\newblock \bibinfo{journal}{\emph{Advances in neural information processing systems}}  \bibinfo{volume}{33} (\bibinfo{year}{2020}), \bibinfo{pages}{18661--18673}.
\newblock


\bibitem[Kohler and Lucchi(2017)]%
        {kohler2017sub}
\bibfield{author}{\bibinfo{person}{Jonas~Moritz Kohler} {and} \bibinfo{person}{Aurelien Lucchi}.} \bibinfo{year}{2017}\natexlab{}.
\newblock \showarticletitle{Sub-sampled cubic regularization for non-convex optimization}. In \bibinfo{booktitle}{\emph{International Conference on Machine Learning}}. PMLR, \bibinfo{pages}{1895--1904}.
\newblock


\bibitem[Kumar et~al\mbox{.}(2017)]%
        {Kumar17}
\bibfield{author}{\bibinfo{person}{B.~G.~V. Kumar}, \bibinfo{person}{B. Harwood}, \bibinfo{person}{G. Carneiro}, \bibinfo{person}{I.~D. Ian D.~Reid}, {and} \bibinfo{person}{T. Drummond}.} \bibinfo{year}{2017}\natexlab{}.
\newblock \showarticletitle{{Smart Mining for Deep Metric Learning}}. In \bibinfo{booktitle}{\emph{ICCV}}.
\newblock


\bibitem[Lee et~al\mbox{.}(2019)]%
        {Lee19}
\bibfield{author}{\bibinfo{person}{K. Lee}, \bibinfo{person}{M.-W. Chang}, {and} \bibinfo{person}{K. Toutanova}.} \bibinfo{year}{2019}\natexlab{}.
\newblock \showarticletitle{Latent retrieval for weakly supervised open domain question answering}. In \bibinfo{booktitle}{\emph{ACL}}.
\newblock


\bibitem[Lindgren et~al\mbox{.}(2021)]%
        {lindgren2021efficient}
\bibfield{author}{\bibinfo{person}{Erik Lindgren}, \bibinfo{person}{Sashank Reddi}, \bibinfo{person}{Ruiqi Guo}, {and} \bibinfo{person}{Sanjiv Kumar}.} \bibinfo{year}{2021}\natexlab{}.
\newblock \showarticletitle{Efficient training of retrieval models using negative cache}.
\newblock \bibinfo{journal}{\emph{Advances in Neural Information Processing Systems}}  \bibinfo{volume}{34} (\bibinfo{year}{2021}), \bibinfo{pages}{4134--4146}.
\newblock


\bibitem[Mittal et~al\mbox{.}(2021a)]%
        {Mittal21}
\bibfield{author}{\bibinfo{person}{A. Mittal}, \bibinfo{person}{K. Dahiya}, \bibinfo{person}{S. Agrawal}, \bibinfo{person}{D. Saini}, \bibinfo{person}{S. Agarwal}, \bibinfo{person}{P. Kar}, {and} \bibinfo{person}{M. Varma}.} \bibinfo{year}{2021}\natexlab{a}.
\newblock \showarticletitle{{DECAF: Deep Extreme Classification with Label Features}}. In \bibinfo{booktitle}{\emph{WSDM}}.
\newblock


\bibitem[Mittal et~al\mbox{.}(2021b)]%
        {Mittal21b}
\bibfield{author}{\bibinfo{person}{A. Mittal}, \bibinfo{person}{N. Sachdeva}, \bibinfo{person}{S. Agrawal}, \bibinfo{person}{S. Agarwal}, \bibinfo{person}{P. Kar}, {and} \bibinfo{person}{M. Varma}.} \bibinfo{year}{2021}\natexlab{b}.
\newblock \showarticletitle{{ECLARE: Extreme Classification with Label Graph Correlations}}. In \bibinfo{booktitle}{\emph{WWW}}.
\newblock


\bibitem[Paszke et~al\mbox{.}(2019)]%
        {paszke2019pytorch}
\bibfield{author}{\bibinfo{person}{Adam Paszke}, \bibinfo{person}{Sam Gross}, \bibinfo{person}{Francisco Massa}, \bibinfo{person}{Adam Lerer}, \bibinfo{person}{James Bradbury}, \bibinfo{person}{Gregory Chanan}, \bibinfo{person}{Trevor Killeen}, \bibinfo{person}{Zeming Lin}, \bibinfo{person}{Natalia Gimelshein}, \bibinfo{person}{Luca Antiga}, \bibinfo{person}{Alban Desmaison}, \bibinfo{person}{Andreas Köpf}, \bibinfo{person}{Edward Yang}, \bibinfo{person}{Zach DeVito}, \bibinfo{person}{Martin Raison}, \bibinfo{person}{Alykhan Tejani}, \bibinfo{person}{Sasank Chilamkurthy}, \bibinfo{person}{Benoit Steiner}, \bibinfo{person}{Lu Fang}, \bibinfo{person}{Junjie Bai}, {and} \bibinfo{person}{Soumith Chintala}.} \bibinfo{year}{2019}\natexlab{}.
\newblock \bibinfo{title}{PyTorch: An Imperative Style, High-Performance Deep Learning Library}.
\newblock
\newblock
\showeprint[arxiv]{1912.01703}~[cs.LG]


\bibitem[Prabhu et~al\mbox{.}(2018)]%
        {Prabhu18b}
\bibfield{author}{\bibinfo{person}{Y. Prabhu}, \bibinfo{person}{A. Kag}, \bibinfo{person}{S. Harsola}, \bibinfo{person}{R. Agrawal}, {and} \bibinfo{person}{M. Varma}.} \bibinfo{year}{2018}\natexlab{}.
\newblock \showarticletitle{{Parabel: Partitioned label trees for extreme classification with application to dynamic search advertising}}. In \bibinfo{booktitle}{\emph{WWW}}.
\newblock


\bibitem[Qaraei and Babbar(2023)]%
        {10.1007/s10994-023-06468-w}
\bibfield{author}{\bibinfo{person}{Mohammadreza Qaraei} {and} \bibinfo{person}{Rohit Babbar}.} \bibinfo{year}{2023}\natexlab{}.
\newblock \showarticletitle{Meta-classifier free negative sampling for extreme multilabel classification}.
\newblock \bibinfo{journal}{\emph{Mach. Learn.}} \bibinfo{volume}{113}, \bibinfo{number}{2} (\bibinfo{date}{nov} \bibinfo{year}{2023}), \bibinfo{pages}{675–697}.
\newblock
\showISSN{0885-6125}
\urldef\tempurl%
\url{https://doi.org/10.1007/s10994-023-06468-w}
\showDOI{\tempurl}


\bibitem[Qaraei et~al\mbox{.}(2021)]%
        {Qaraei21}
\bibfield{author}{\bibinfo{person}{M. Qaraei}, \bibinfo{person}{E. Schultheis}, \bibinfo{person}{P. Gupta}, {and} \bibinfo{person}{R. Babbar}.} \bibinfo{year}{2021}\natexlab{}.
\newblock \showarticletitle{Convex Surrogates for Unbiased Loss Functions in Extreme Classification With Missing Labels}. In \bibinfo{booktitle}{\emph{The WebConf}}.
\newblock


\bibitem[Qu et~al\mbox{.}(2021)]%
        {rocketqa}
\bibfield{author}{\bibinfo{person}{Yingqi Qu}, \bibinfo{person}{Yuchen Ding}, \bibinfo{person}{Jing Liu}, \bibinfo{person}{Kai Liu}, \bibinfo{person}{Ruiyang Ren}, \bibinfo{person}{Wayne~Xin Zhao}, \bibinfo{person}{Daxiang Dong}, \bibinfo{person}{Hua Wu}, {and} \bibinfo{person}{Haifeng Wang}.} \bibinfo{year}{2021}\natexlab{}.
\newblock \showarticletitle{RocketQA: An Optimized Training Approach to Dense Passage Retrieval for Open-Domain Question Answering}. In \bibinfo{booktitle}{\emph{Proceedings of the 2021 Conference of the North American Chapter of the Association for Computational Linguistics: Human Language Technologies}}. \bibinfo{pages}{5835--5847}.
\newblock


\bibitem[Reddi et~al\mbox{.}(2019)]%
        {Reddi19}
\bibfield{author}{\bibinfo{person}{S.~J. Reddi}, \bibinfo{person}{S. Kale}, \bibinfo{person}{F.X. Yu}, \bibinfo{person}{D.~N.~H. Rice}, \bibinfo{person}{J. Chen}, {and} \bibinfo{person}{S. Kumar}.} \bibinfo{year}{2019}\natexlab{}.
\newblock \showarticletitle{{Stochastic Negative Mining for Learning with Large Output Spaces}}. In \bibinfo{booktitle}{\emph{AISTATS}}.
\newblock


\bibitem[Saini et~al\mbox{.}(2021)]%
        {Saini21}
\bibfield{author}{\bibinfo{person}{D. Saini}, \bibinfo{person}{A.K. Jain}, \bibinfo{person}{K. Dave}, \bibinfo{person}{J. Jiao}, \bibinfo{person}{A. Singh}, \bibinfo{person}{R. Zhang}, {and} \bibinfo{person}{M. Varma}.} \bibinfo{year}{2021}\natexlab{}.
\newblock \showarticletitle{{GalaXC: Graph Neural Networks with Labelwise Attention for Extreme Classification}}. In \bibinfo{booktitle}{\emph{WWW}}.
\newblock


\bibitem[Sanh et~al\mbox{.}(2019)]%
        {Sanh2019DistilBERTAD}
\bibfield{author}{\bibinfo{person}{V. Sanh}, \bibinfo{person}{L. Debut}, \bibinfo{person}{J. Chaumond}, {and} \bibinfo{person}{T. Wolf}.} \bibinfo{year}{2019}\natexlab{}.
\newblock \showarticletitle{{DistilBERT, a distilled version of BERT: smaller, faster, cheaper and lighter}}.
\newblock \bibinfo{journal}{\emph{ArXiv}} (\bibinfo{year}{2019}).
\newblock


\bibitem[Schultheis et~al\mbox{.}(2022)]%
        {Schultheis22}
\bibfield{author}{\bibinfo{person}{E. Schultheis}, \bibinfo{person}{M. Wydmuch}, \bibinfo{person}{R. Babbar}, {and} \bibinfo{person}{K. Dembczynski}.} \bibinfo{year}{2022}\natexlab{}.
\newblock \showarticletitle{On Missing Labels, Long-Tails and Propensities in Extreme Multi-Label Classification}. In \bibinfo{booktitle}{\emph{KDD}}.
\newblock


\bibitem[Simhadri et~al\mbox{.}(2023)]%
        {diskann-github}
\bibfield{author}{\bibinfo{person}{Harsha~Vardhan Simhadri}, \bibinfo{person}{Ravishankar Krishnaswamy}, \bibinfo{person}{Gopal Srinivasa}, \bibinfo{person}{Suhas~Jayaram Subramanya}, \bibinfo{person}{Andrija Antonijevic}, \bibinfo{person}{Dax Pryce}, \bibinfo{person}{David Kaczynski}, \bibinfo{person}{Shane Williams}, \bibinfo{person}{Siddarth Gollapudi}, \bibinfo{person}{Varun Sivashankar}, \bibinfo{person}{Neel Karia}, \bibinfo{person}{Aditi Singh}, \bibinfo{person}{Shikhar Jaiswal}, \bibinfo{person}{Neelam Mahapatro}, \bibinfo{person}{Philip Adams}, \bibinfo{person}{Bryan Tower}, {and} \bibinfo{person}{Yash Patel}.} \bibinfo{year}{2023}\natexlab{}.
\newblock \bibinfo{title}{{DiskANN: Graph-structured Indices for Scalable, Fast, Fresh and Filtered Approximate Nearest Neighbor Search}}.
\newblock
\newblock
\urldef\tempurl%
\url{https://github.com/Microsoft/DiskANN}
\showURL{%
\tempurl}


\bibitem[Subramanya et~al\mbox{.}(2019)]%
        {Subramanya19}
\bibfield{author}{\bibinfo{person}{S.~J. Subramanya}, \bibinfo{person}{Devvrit}, \bibinfo{person}{R. Kadekodi}, \bibinfo{person}{R. Krishnaswamy}, {and} \bibinfo{person}{H. Simhadri}.} \bibinfo{year}{2019}\natexlab{}.
\newblock \showarticletitle{{DiskANN: Fast Accurate Billion-point Nearest Neighbor Search on a Single Node}}. In \bibinfo{booktitle}{\emph{NeurIPS}}.
\newblock


\bibitem[Xiong et~al\mbox{.}(2020)]%
        {xiong2020approximate}
\bibfield{author}{\bibinfo{person}{Lee Xiong}, \bibinfo{person}{Chenyan Xiong}, \bibinfo{person}{Ye Li}, \bibinfo{person}{Kwok-Fung Tang}, \bibinfo{person}{Jialin Liu}, \bibinfo{person}{Paul~N Bennett}, \bibinfo{person}{Junaid Ahmed}, {and} \bibinfo{person}{Arnold Overwijk}.} \bibinfo{year}{2020}\natexlab{}.
\newblock \showarticletitle{Approximate Nearest Neighbor Negative Contrastive Learning for Dense Text Retrieval}. In \bibinfo{booktitle}{\emph{International Conference on Learning Representations}}.
\newblock


\bibitem[Zhai et~al\mbox{.}(2023)]%
        {zhai2023sigmoid}
\bibfield{author}{\bibinfo{person}{Xiaohua Zhai}, \bibinfo{person}{Basil Mustafa}, \bibinfo{person}{Alexander Kolesnikov}, {and} \bibinfo{person}{Lucas Beyer}.} \bibinfo{year}{2023}\natexlab{}.
\newblock \showarticletitle{Sigmoid Loss for Language Image Pre-Training}.
\newblock \bibinfo{journal}{\emph{arXiv e-prints}} (\bibinfo{year}{2023}), \bibinfo{pages}{arXiv--2303}.
\newblock


\end{thebibliography}

  
\appendix
\onecolumn

\section{Proofs}
\label{sec:proofs}
\subsection{Proof of Lemma \ref{lem:estimator}}
\textbf{(1)} The first part of the Lemma is easy to show. Note that the expectation is with respect to the randomness in the sampling of negative labels with uniform probability $p = 1/(L - \hardnegsize)$ from the set $[L] \setminus \hardneg$. Comparing the full loss \eqref{eqn:reneeloss} and the estimator in \eqref{eqn:lexusloss}, it suffices to show that $\loss_-(\encparam, \W; \x, \hardneg) + \frac{1}{p \cdot \randnegsize} \mathbf{E}[\loss_-(\encparam, \W; \randneg)] = \loss_-(\encparam, \W; \x; [L])$. This is true because $\mathbf{E}[\loss_-(\encparam, \W; \x, \randneg)] = \\ \frac{\randnegsize}{(L - \hardnegsize)} \sum_{\ell \in [L] \setminus \hardneg} \loss_-(\encparam, \W; \x, \{\ell\}) = \frac{k_r}{(L-k_h)}\loss_-(\encparam, \W; \x, [L] \setminus \hardneg) = p\cdot k_r\loss_-(\encparam, \W; \x, [L] \setminus \hardneg)$. The same argument holds for gradients as well.

\textbf{(2)} Now consider the second part of the lemma, which is about concentration of gradients around its expectation. For ease of presentation, we will work with mini-batch of size 1, i.e., the query data-point $\x$, and we will elide the arguments from the loss functions using ``$\cdot$" when it's clear from the context. 

\textbf{(2.1)} First, let us consider the gradient of $\loss_\lexus$ with respect to $\encparam$. Our goal is to bound the likelihood of the bad event that $\|\nabla_{\encparam} \loss_\lexus ( \cdot\ 
 ; \x) - \nabla_{\encparam} \loss ( \cdot\ ; \x)\| \geq \epsilon \| \nabla_{\encparam} \loss ( \cdot\ ; \x)\|$, for a given $\epsilon > 0$. 
 
 Firstly, recall from the first part of the proof that the losses $\loss_\lexus$ and $\loss$, and therefore gradients, differ exactly in the part of the loss function where we do random negative sampling, and the remainder is identical across the two loss functions. Thus, 
 \begin{equation}
     \|\nabla_{\encparam} \loss_\lexus ( \cdot\ ; \x) - \nabla_{\encparam} \loss ( \cdot\ ; \x)\| =  \bigg\|\frac{1}{p \cdot \randnegsize}\nabla_{\encparam} \loss_- ( \cdot\ 
 ; \x, \randneg) - \nabla_{\encparam} \loss_{-}( \cdot\ ; \x, [L] \setminus \hardneg)\bigg\|.
 \label{eqn:inter1}
 \end{equation}
 Now let $$\hat{Z}_{\ell} = \frac{1}{p} \nabla_{\encparam}\loss_- ( \cdot\ ; \x, \{\ell\}),$$ and let $$Z := \nabla_{\encparam}\loss_{-}( \cdot\ ; \x, [L] \setminus \hardneg).$$
 Now, we can rewrite RHS of \eqref{eqn:inter1} as:
 \begin{equation}
     \|\nabla_{\encparam} \loss_\lexus ( \cdot\ ; \x) - \nabla_{\encparam} \loss ( \cdot\ ; \x)\| = \bigg\|\frac{1}{\randnegsize}\sum_{\ell \in \randneg} \hat{Z}_{\ell} - Z\bigg\|.
 \label{eqn:inter2}
 \end{equation}
 Note that since $\mathbb{E}[\hat{Z}_{\ell}] = Z$, we have average of independent, vector-valued, zero-mean random variables in \eqref{eqn:inter2}. Further, we can bound $\mathbb{E} [\|\hat{Z}_{\ell}\|] = \frac{1}{p} E\bigg[\bigg\|\frac{-1}{1 + \exp(-\w_\ell^T \phi(\x))}\w_\ell\nabla_{\encparam}\phi(\x)\bigg\|\bigg] \leq \frac{1}{p}\|\w_\ell\|\|\nabla_{\encparam}\phi(\x)\| \leq \mu = \mathcal{O}(L)$, where the hidden constants depend on the Lipschitz-constant of the encoder function $\enc(\x)$, and the (bounded) norm of the classifiers. Similarly we can bound the second moment as $\mathbb{E} [\|\hat{Z}_{\ell}\|^2] \leq \sigma^2 = \mathcal{O}(L^2)$. 
 
Now we can appeal to the vector version of Bernstein's inequality as given in Lemma 18 of \citet{kohler2017sub}, Appendix A, as all the necessary quantities satisfy the pre-requisites of the Lemma. We have, for $0 < \epsilon' < \sigma^2/\mu$, $$P\bigg(\bigg\|\frac{1}{\randnegsize}\sum_{\ell \in \randneg} \hat{Z}_{\ell} - Z\bigg\| \geq \epsilon'\bigg) \leq 2 \exp\bigg(-\randnegsize \cdot \frac{\epsilon'^2}{8\sigma^2}\bigg).$$

Now, let's substitute $\epsilon' = \epsilon \|\nabla_{\encparam} \loss ( \cdot\ ; \x)\|$, which is the bound we seek. This is permissible as $\|\nabla_{\encparam} \loss ( \cdot\ ; \x)\| = \mathcal{O}(L)$, as is stipulated by the bound $\epsilon' < \sigma^2/\mu = \mathcal{O}(L)$. Now, setting $\delta \geq 2 \exp\bigg(-\randnegsize \cdot \frac{\epsilon'^2}{8\sigma^2}\bigg)$, and using the fact that $\epsilon'^2/\sigma^2 = \mathcal{O}(1)$, and re-arranging terms, we get that if $\randnegsize \geq \mathcal{O}\big(\log(1/\delta)\frac{1}{\epsilon^2}\big)$, with probability at least $1 - \delta$, \eqref{eqn:inter2} is upper bounded by $\epsilon \| \nabla_{\encparam} \loss ( \cdot\ ; \x)\|$ as desired.

\textbf{(2.2)} Next, let us consider the gradient of $\loss_\lexus$ with respect to classifiers $\W$. Our goal is to bound the likelihood of the bad event that $\|\nabla_{\W} \loss_\lexus ( \cdot\ 
 ; \x) - \nabla_{\W} \loss ( \cdot\ ; \x)\| \geq \epsilon \| \nabla_{\W} \loss ( \cdot\ ; \x)\|$, for a given $\epsilon > 0$. The gradient is a $\embsize \times L$ matrix, where the $\ell$th column has the gradient of the loss w.r.t. $\w_\ell$, and the $\| \cdot \|$ denotes the $L_2$ norm of the vectorized matrix.
Here, we can not apply Bernstein inequality as in the above case. But, we can try to bound the differences directly as follows. 

Denote by $\ghat_\ell$ the $\ell$th column in the gradient matrix for $\loss_\lexus$ and by $\g_\ell$ the $\ell$th column in the gradient matrix for $\loss$. Note that for $\ell \in \randneg$, $\ghat_\ell - \g_\ell = 0$. And $\|\ghat_\ell - \g_\ell\| = \|\g_\ell\|$ for the labels that are not sampled. So,  using the vectorized $L_2$-norm, $$\|\nabla_{\W} \loss_\lexus ( \cdot\ ; \x) - \nabla_{\W} \loss ( \cdot\ ; \x)\|^2 = \sum_{\ell \notin \randneg} \|\g_\ell\|^2.$$ We can apply Chernoff bounds to control the deviation from the true gradient, which gives 

\begin{align}
P\bigg(\|\nabla_{\W} \loss_\lexus ( \cdot\ ; \x) - \nabla_{\W} \loss ( \cdot\ ; \x)\| \geq \epsilon \|\nabla_{\W} \loss ( \cdot\ ; \x)\|\bigg) \\ &\leq \frac{\mathbb{E}\big[ \exp(s\cdot  \|\nabla_{\W} \loss_\lexus ( \cdot\ ; \x) - \nabla_{\W} \loss ( \cdot\ ; \x)\| )\big]}{\exp(s \cdot \epsilon \|\nabla_{\W} \loss ( \cdot\ ; \x)\|)} \\ &= \frac{\mathbb{E}\big[ \exp(s\cdot  \sqrt{\sum_{\ell \notin \randneg} \|\g_\ell\|^2} )\big]}{\exp(s \cdot \epsilon \|\nabla_{\W} \loss ( \cdot\ ; \x)\|)}.
\label{eqn:inter3}
\end{align}

Now consider $\|\g_\ell\| = \bigg\|\frac{1}{1 + \exp(-\w_\ell^T\Phi(\x))}\Phi(\x)\bigg\| \leq c$, where $c$ is a constant the depends on the bound of the embedding and classifier norms. We can bound $\mathbb{E}\big[\exp(s\cdot  \sqrt{\sum_{\ell \notin \randneg} \|\g_\ell\|^2} )\big] \leq \exp(s\cdot  (L - \randnegsize)c )$, so \eqref{eqn:inter3} becomes:

$$ \text{RHS  of } \eqref{eqn:inter3} \leq \exp(s\cdot  (L - \randnegsize)c ) \cdot\ \exp(-s \cdot \epsilon \|\nabla_{\W} \loss ( \cdot\ ; \x)\|) \leq \exp(s\cdot  (L - \randnegsize)c ).$$
We can choose $s > 0$ that optimizes the upper bound, i.e., probability of the bad event that $\|\nabla_{\W} \loss_\lexus ( \cdot\ ; \x) - \nabla_{\W} \loss ( \cdot\ ; \x)\| \geq \epsilon \| \nabla_{\W} \loss ( \cdot\ ; \x)\|$, above. 

This concludes the proof of the Lemma.


\subsection{Proof of Theorem \ref{thm:convergence}}
\label{sec:assumption}
We will state a smoothness assumption on the loss function.
\begin{assumption} The loss $\loss$ is $H$-smooth in its parameters $\encparam, \W$, i.e., for all valid parameters $(\encparam_1, \W_1)$,  $(\encparam_2, \W_2)$, we have:
$$ \loss(\encparam_1, \W_1) \leq \loss(\encparam_2, \W_2) + \langle \nabla \loss(\encparam_2, \W_2), [\encparam_1 - \encparam_2; \W_1 - \W_2] \rangle + \frac{H}{2} \big\|[\encparam_1 - \encparam_2; \W_1 - \W_2] \big\|_2^2 .$$
Lemma \ref{lem:estimator} suggests that during every epoch, Algorithm \ref{algo:LEXUS} has access to a biased gradient oracle, i.e., for some $\epsilon \in (0,1)$, for all epochs $t$, with high probability, we have the vectorized-gradient of the loss with respect to the parameters $\encparam$ and $\W$ at epoch $t$ , say $\g^{(t)}$, can be written as $\g^{(t)} = \nabla \loss(\encparam^{(t)}, \W^{(t)}) + \Delta^{(t)}$, where $\|\Delta^{(t)}\|_2 \leq \epsilon \cdot \|\nabla \loss(\encparam^{(t)}, \W^{(t)})\|_2^2$. If we choose the learning rate $\eta$ as stated in the Theorem, then we satisfy the pre-requisites of Lemma 7 in the Appendix of \cite{Dahiya23}. The convergence guarantee as given in the statement of the Theorem follows, with constants as specified in Lemma 7 of \cite{Dahiya23}. This concludes the proof.
\label{assumption}
\end{assumption}

\section{Implementation Details}
\label{sec:implementation_app}
\subsection{Datasets}
\label{sec:datasets_app}
In this subsection, we provide more details on the datasets used for our experiments. We compare \lexus{} on short-text datasets such as LF-AmazonTitles-131K, LF-AmazonTitles-1.3M and long-text datasets such as LF-Amazon-131K and LF-Wikipedia-500K. All these four datasets are with label features. We also evaluate on Amazon-670K and Amazon-3M datasets which are without label features.  We also report results on proprietary datasets with $20$ million and $120$ million labels. These datasets cover a variety of applications including product-to-product recommendation (Amazon-670K, Amazon-3M, LF-Amazon-131K, LF-AmazonTitles-131K, and LF-AmazonTitles-1.3M), predicting Wikipedia categories (LF-Wikipedia-500K) and matching user queries to advertiser bid phrases in sponsored search (SS-20M, SS-120M)Table~\ref{tab:datastats} in the Appendix gives statistic of these datasets. In addition to the aggregate statistics in Table~\ref{tab:datastats}, we did a decile-wise analysis of SS-120M and LF-AmazonTitles-131K to compare the characterisitcs of the proprietary dataset with academic dataset. We divided the labels into 10 deciles based on number of training points and calculate the number of labels in each decile to show that the proprietary dataset has similar distribution of head and tail labels. Table~\ref{tab:decile} in the Appendix has the decile-wise analysis. Similar to \renee, we augment the training data with label texts as training data-points with the corresponding label id as a positive label.

\begin{table*}[]
\caption{Dataset statistics.  All the public datasets can be downloaded from the XC repository~\cite{XMLRepo}.} 
 \label{tab:datastats}
\centering
\resizebox{\textwidth}{!}{
\begin{tabular}{clrrrrr}
\hline
\textbf{Dataset}     & \textbf{Type} & \multicolumn{1}{c}{\textbf{\begin{tabular}[c]{@{}c@{}}\# Training\\  Points\end{tabular}}} & \multicolumn{1}{c}{\textbf{\# Labels}} & \multicolumn{1}{c}{\textbf{\# Test points}} & \multicolumn{1}{c}{\textbf{\begin{tabular}[c]{@{}c@{}}Avg. \# \\ datapts.\\  per label\end{tabular}}} & \multicolumn{1}{c}{\textbf{\begin{tabular}[c]{@{}c@{}}Avg. \#\\ labels per \\ datapt.\end{tabular}}} \\ \hline
\multicolumn{7}{c}{Datasets without label features}                                                                                                                                                                                                                                                                                                                                                                                              \\ \hline
Amazon-670K          & Full-text     & 490,449                                                                                    & 670,091                                & 153,025                                     & 3.9                                                                                                   & 5.5                                                                                                  \\
Amazon-3M            & Full-text     & 1,717,899                                                                                  & 2,812,281                              & 742,507                                     & 22.0                                                                                                  & 36.0                                                                                                 \\ \hline
\multicolumn{7}{c}{Datasets with label features}                                                                                                                                                                                                                                                                                                                                                                                                 \\ \hline
LF-AmazonTitles-131K & Short-text    & 294,805                                                                                    & 131,073                                & 134,835                                     & 2.3                                                                                                   & 5.2                                                                                                  \\
LF-AmazonTitles-1.3M & Short-text    & 2,248,619                                                                                  & 1,305,265                              & 970,237                                     & 22.2                                                                                                  & 38.2                                                                                                 \\
LF-Amazon-131K       & Full-text     & 294,805                                                                                    & 131,073                                & 134,835                                     & 2.3                                                                                                   & 5.2                                                                                                  \\
LF-Wikipedia-500K    & Full-text     & 1,813,391                                                                                  & 501,070                                & 783,743                                     & 4.8                                                                                                   & 24.8                                                                                                 \\ \hline
\multicolumn{7}{c}{Proprietary Datasets}                                                                                                                                                                                                                                                                                                                                                                                                \\ \hline
SS-20M             & Short-text    & 146,193,053                                                                                & 19,049,952                             & 16,629,030                                  & 199.7                                                                                                 & 26.0                                                                                                 \\
SS-120M            & Short-text    & 370,080,440                                                                                & 120,293,341                            & 92,532,582                                  & 10.4                                                                                                  & 10.4                                                                                                 \\ \hline
\end{tabular}
}
\end{table*}

\begin{table*}[]

\caption{This table contains the \% of labels in each decile for LF-AmazonTitles-131K and SS-120M datasets and show that their head and tail distributions are similar} 
 \label{tab:decile}

\centering
 \resizebox{\textwidth}{!}{

    \begin{tabular}{@{}ll|l|l|l|l|l|l|l|l|l|l@{}}
    \toprule
    \multicolumn{2}{c|}{\textbf{Decile}} & 10 & 9 & 8 & 7 & 6 & 5 & 4 & 3 & 2 & 1 \\ \midrule
    \multicolumn{2}{l|}{\textbf{LF-AmazonTitles-131K}} & 0.3\% & 1.0\% & 1.8\% & 2.8\% & 4.3\% & 6.4\% & 9.4\% & 14.0\% & 21.5\% & 38.5\% \\ \midrule
    \multicolumn{2}{l|}{\textbf{SS-120M}} & 1.2\% & 3.1\% & 4.5\% & 5.9\% & 7.3\% & 8.9\% & 10.9\% & 13.5\% & 17.8\% & 26.9\% \\ \bottomrule
    \end{tabular}

}

\end{table*}

\subsection{Hyperparameter Tuning}
\label{sec:hyperparamtuning}
For \lexus{}, we use the same values as that of \renee{} for hyperparameters batch-size and weight decay. We tune dropout in the range [0.5-0.9] on a small validation set and observe that higher dropout performs better across datasets. We tune learning rate for \lexus{} in range [10 * \lr\_\renee, 0.1 * \lr\_\renee] for both encoder and classifier on a small validation set. In general, we observe that higher $lr$ values yield good performance. Similarly, we tune warmup in range [2000-10000] on a small validation set and observe that lower warmup values help perform better. As \lexus{} achieves convergence faster than \renee{}, the number of epochs for \lexus{} is smaller than \renee{} for many datasets.

There are three hyperparameters for negative sampling: $\positivesize$, $\randnegsize$ and $\hardnegsize$.
$\positivesize$: For all public datasets, we simply use all the available +ve labels for the data points. e.g., $\positivesize$ = 7, which is the max. no. of +ve labels for any data point, for LF-AmazonTitles-131K; in case a data point has fewer positives than 7, we simply pad that with random negatives. 

$\randnegsize$, $\hardnegsize$: We tune $\randnegsize$ in (1600,3200,4800,6400) and $\hardnegsize$ in (160,240,320,400,480,560,640,720,800) on a small validation set and pick the one that works best. In general, we have observed that the skewed ratio $\randnegsize$:$\hardnegsize$ in range(10:1 to 5:1) works better across datasets.

We tuned $\taus$ in (5,10,20,40) and $\taur$ in range (5,10,20) for 2 datasets, LF-AmazonTitles-131K and LF-AmazonTitles-1.3M. In general setting both the values to 5 seems to work the best, hence we use the same value for all other datasets.

\begin{table*}[]
\caption{Hyperparameter values of \lexus{} to facilitate reproducibility. The batch-size, $k_p$, $k_h$ and $k_r$ values are per GPU, when run on 16 GPUs. The ANNS index start epoch $\tau_s$ and refresh interval $\tau_r$ are both set as 5 for all the datasets.} 
 \label{tab:hyperparam}
\centering
\resizebox{\textwidth}{!}{
\begin{tabular}{c|rlrrrlrrrr}
\hline
\textbf{Dataset}     & \multicolumn{1}{c}{\textbf{batch-size}} & \multicolumn{1}{c}{\textbf{epochs}} & \multicolumn{1}{c}{\textbf{\begin{tabular}[c]{@{}c@{}}lr\\ (encoder)\end{tabular}}} & \multicolumn{1}{c}{\textbf{\begin{tabular}[c]{@{}c@{}}lr\\ (classifier)\end{tabular}}} & \multicolumn{1}{c}{\textbf{dropout}} & \multicolumn{1}{c}{\textbf{warmup}} & \multicolumn{1}{c}{\textbf{\begin{tabular}[c]{@{}c@{}}weight decay\\ (classifier)\end{tabular}}} & \multicolumn{1}{c}{\textbf{$k_r$}} & \multicolumn{1}{c}{\textbf{$k_h$}} & \multicolumn{1}{c}{\textbf{$k_p$}} \\ \hline
LF-AmazonTitles-131K & 32                                     & 75                                 & 0.00001                                                                             & 0.05                                                                                   & 0.85                                 & 3000                                & 0.0001                                                                                          & 400                               & 50                               & 7                                 \\
LF-Amazon-131K       & 32                                     & 100                                 & 0.0001                                                                              & 0.05                                                                                   & 0.75                                 & 5000                                & 0.0001                                                                                          & 100                               & 15                                & 7                                 \\
LF-Wikipedia-500K       & 128                                     & 40                                 & 0.0001                                                                              & 0.002                                                                                   & 0.7                                 & 2500                                & 0.001                                                                                          & 200                               & 30                                & 13                                 \\
LF-AmazonTitles-1.3M & 64                                    & 100                                 & 0.000003                                                                            & 0.01                                                                                   & 0.75                                 & 5000                                & 0.0001                                                                                          & 100                               & 15                                & 79                                 \\
Amazon-670K          & 16                                     & 70                                  & 0.00004                                                                             & 0.01                                                                                   & 0.8                                  & 10000                               & 0.001                                                                                           & 400                                & 50                                & 7                               \\
Amazon-3M            & 16                                     & 60                                  & 0.00004                                                                             & 0.01                                                                                   & 0.75                                 & 10000                               & 0.001                                                                                           & 400                               & 50                                & 50                                \\ \hline
\end{tabular}
}
\end{table*}

\subsection{Break-up of iteration time}
\label{sec:breakup_appendix}
Table \ref{tab:timebreakup} shows the break-up of iteration time for three datasets. The different steps involved in an iteration are - 
\begin{itemize}
    \item  Data Preparation: This includes creating a batch of data-points along with their positives and negatives. This is done on CPU when the training is going on GPU.
    \item CopyDataToGPU: This involves copying prepared data to GPU.
    \item EncFwdPass: This refers to the encoder forward pass where the input data-points are passed through the encoder to generate the embeddings.
    \item ClfFwdPass: This refers to the classifier forward pass which involves matrix multiplication $O(L)$ in \renee. This becomes the main bottleneck as $L$ increases. \lexus{} helps reduce this matrix multiplication to $O(log(L))$.
    \item BwdPass: This refers to the backward pass of both the encoder and the classifier. 
\end{itemize}

\subsection{Data and Model Parallel Training}
\label{sec:implemenation_appendix}
As mentioned in Section \ref{sec:algorithm}, \lexus{} implements data-parallel training for encoder and model-parallel training for classifier in multi-GPU setting. Consider a scenario with $G$ GPUs, $L$ labels, and $B$ batch size. In multi-GPU setting, the encoder will produce the embeddings of the input queries in parallel, with each GPU processing $\frac{B}{G}$ input queries. An all-gather call is then used to distribute the embeddings to all GPUs to attain the classifiers. The classifiers are divided across GPUs, each GPU processing $\frac{L}{G}$ classifiers. The model-parallel approach helps in speeding up the hard-negative retrieval process as well. For very large datasets, instead of building ANNS index on $L$ labels, multiple ANNS indices are built on, one per GPU using $\frac{L}{G}$ labels present on that GPU.
To further optimize memory and compute, we use a bottleneck layer that reduces the classifier dimension to 64 for large datasets similar to \renee. 


\section{Additional Results}
This section additional results for all the datasets which were not covered in the main paper.

\label{sec:addresults}

\begin{table*} []
 \caption{Detailed results on short-text datasets. TT refers to training time in hours on a single Nvidia V100 GPU.}
    \label{tab:supp:results_short_text}
      \centering
      \resizebox{\linewidth}{!}{
        \begin{tabular}{@{}l|ccccccccccc@{}}
        \toprule
        \textbf{Method} & \textbf{P@1} & \textbf{P@3} & \textbf{P@5} & \textbf{N@3} & \textbf{N@5} & \textbf{PSP@1} & \textbf{PSP@3} & \textbf{PSP@5} & \textbf{PSN@3} & \textbf{PSN@5} & \textbf{TT} \\ \midrule
        
        \midrule
        \multicolumn{12}{c}{LF-AmazonTitles-1.3M}\\ \midrule
        \lexus{} & 55.71 & 48.95 & 44.26 & 53.22 & 52.06 & 30.23 & 34.42 & 36.84 & 33.36 & 35.27 & 24.86 \\
        \renee{} & 56.04 & 49.91 & 45.32 & 54.21 & 53.15 & 28.54 & 33.38 & 36.14 & 32.15 & 34.18 & 27.33 \\
        \dexa & 55.76 & 48.07 & 42.95 & 52.81 & 51.34 & 30.01 & 33.37 & 35.29 & 32.7 & 34.33 & 103.13 \\
        \dexa{} (ensemble) &  56.63 & 49.05 & 43.90 & 53.81 & 52.37 & 29.12 & 32.69 & 34.86 & 32.02 & 33.86 & 103.13 \\
        NGAME & 54.69 & 47.76 & 42.8 & 52.21 & 50.85 & 28.23 & 32.26 & 34.48 & 31.29 & 33.04 & 97.75\\
        NGAME (ensemble) & 56.75 & 49.19 & 44.09 & 53.84 & 52.41 & 29.18 & 33.01 & 35.36 & 32.07 & 33.91 &  97.75 \\
        \dexml{} & 58.40 & - & 45.46 & - & 54.30 & - & - & 36.58 & - & - & $\sim$2K  \\
        SiameseXML & 49.02 & 42.72 & 38.52 & 46.38 & 45.15 & 27.12	& 30.43 & 32.52 & 29.41 & 30.9 & 9.89 \\ 
        ECLARE & 50.14 & 44.09 & 40 & 47.75 & 46.68 & 23.43 & 27.9 & 30.56 & 26.67 & 28.61 & 70.59 \\ 
        GalaXC & 49.81 & 44.23 & 40.12 & 47.64 & 46.47 & 25.22 & 29.12 & 31.44 & 27.81 & 29.36 & 9.55 \\ 
        DECAF & 50.67 & 44.49 & 40.35 & 48.05 & 46.85 & 22.07 & 26.54 & 29.3 & 25.06 & 26.85 & 74.47 \\ 
        Astec & 48.82 & 42.62 & 38.44 & 46.11 & 44.8 & 21.47 & 25.41 & 27.86 & 24.08 & 25.66 & 18.54 \\ 
        AttentionXML & 45.04 & 39.71 & 36.25 & 42.42 & 41.23 & 15.97 & 19.9 & 22.54 & 18.23 & 19.6 & 380.02 \\ 
        MACH & 35.68 & 31.22 & 28.35 & 33.42 & 32.27 & 9.32 & 11.65 & 13.26 & 10.79 & 11.65 & 60.39 \\ 
        X-Transformer & - & - & - & - & - & - & - & - & - & - & - \\ 
        LightXML & - & - & - & - & - & - & - & - & - & - & - \\ 
        AnneXML & 47.79 & 41.65 & 36.91 & 44.83 & 42.93 & 15.42 & 19.67 & 21.91 & 18.05 & 19.36 & 2.48 \\ 
        DiSMEC & - & - & - & - & - & - & - & - & - & - & - \\ 
        Parabel & 46.79 & 41.36 & 37.65 & 44.39 & 43.25 & 16.94 & 21.31 & 24.13 & 19.7 & 21.34 & 1.5 \\ 
        XT & 40.6 & 35.74 & 32.01 & 38.18 & 36.68 & 13.67 & 17.11 & 19.06 & 15.64 & 16.65 & 82.18 \\ 
        Slice & 34.8 & 30.58 & 27.71 & 32.72 & 31.69 & 13.96 & 17.08 & 19.14 & 15.83 & 16.97 & 0.79 \\ 
        PfastreXML & 37.08 & 33.77 & 31.43 & 36.61 & 36.61 & 28.71 & 30.98 & 32.51 & 29.92 & 30.73 & 9.66 \\ 
        Bonsai & 47.87 & 42.19 & 38.34 & 45.47 & 44.35 & 18.48 & 23.06 & 25.95 & 21.52 & 23.33 & 7.89 \\ 
        XR-Transformer & 50.14 & 44.07 & 39.98 & 47.71 & 46.59 & 20.06 & 24.85 & 27.79 & 23.44 & 25.41 & 132 \\ 

        \midrule
        \multicolumn{12}{c}{LF-AmazonTitles-131K}\\ \midrule
        \lexus{} & 46.20 & 30.72 & 21.95 &  47.43 & 49.57 & 39.36& 47.43 & 50.57 & 43.81 & 46.46 & 2.02 \\
        \renee{} &  46.05& 30.81& 22.04&  47.46& 49.68& 39.08& 45.12& 50.48 &43.56& 46.24   &  2.05\\
        \dexa & 45.78 & 30.13 & 21.29 &   46.49 & 48.37 & 38.57 & 43.95 & 48.56 &   42.44 & 44.76  & 13.01 \\ 
        \dexa{} (ensemble) & 46.42 & 30.50 & 21.59 & 47.06 & 49.00 & 39.11 & 44.69 & 49.65 & 43.10 & 45.58 & 13.01 \\ 
        NGAME & 44.95 & 29.87 & 21.20 & 45.95 & 47.92 & 38.25 & 43.75 & 48.42 & 42.18 & 44.53 & 12.59 \\
        NGAME (ensemble) & 46.01 & 30.28 & 21.47 & 46.69 & 48.67 & 38.81 & 44.4 & 49.43 & 42.79 & 45.31 & 12.59 \\
        \dexml{} & 42.52 & - & 20.64 & - & 46.33 & - & - & 47.40 & - & - & 83.33 \\
        SiameseXML & 41.42 & 27.92 & 21.21 & 42.65 & 44.95 & 35.80 & 40.96 & 46.19 & 39.36 & 41.95 & 1.08 \\ 
        ECLARE & 40.74 & 27.54 & 19.88 & 42.01 & 44.16 & 33.51 & 39.55 & 44.7 & 37.7 & 40.21 & 2.16 \\ 
        GalaXC & 39.17 & 26.85 & 19.49 & 40.82 & 43.06 & 32.5 & 38.79 & 43.95 & 36.86 & 39.37 & 0.42 \\ 
        DECAF & 38.4 & 25.84 & 18.65 & 39.43 & 41.46 & 30.85 & 36.44 & 41.42 & 34.69 & 37.13 & 2.16 \\ 
        Astec & 37.12 & 25.2 & 18.24 & 38.17 & 40.16 & 29.22 & 34.64 & 39.49 & 32.73 & 35.03 & 1.83 \\ 
        AttentionXML & 32.25 & 21.7 & 15.61 & 32.83 & 34.42 & 23.97 & 28.6 & 32.57 & 26.88 & 28.75 & 20.73 \\ 
        MACH & 33.49 & 22.71 & 16.45 & 34.36 & 36.16 & 24.97 & 30.23 & 34.72 & 28.41 & 30.54 & 3.3 \\ 
        X-Transformer & 29.95 & 18.73 & 13.07 & 28.75 & 29.6 & 21.72 & 24.42 & 27.09 & 23.18 & 24.39 & 64.4 \\ 
        LightXML & 35.6 & 24.15 & 17.45 & 36.33 & 38.17 & 25.67 & 31.66 & 36.44 & 29.43 & 31.68 & 71.4 \\ 
        BERTXML & 38.89	& 26.17	& 18.72	& 39.93	& 41.79	& 30.1	& 36.81 &	41.85 &	34.8 & 37.28 & 12.55\\
        ELIAS & 40.13 & 27.11 & 19.54 & - & - & 31.05 & 37.57 & 42.88 & - & - & -\\
        AnneXML & 30.05 & 21.25 & 16.02 & 31.58 & 34.05 & 19.23 & 26.09 & 32.26 & 23.64 & 26.6 & 0.08 \\ 
        DiSMEC & 35.14 & 23.88 & 17.24 & 36.17 & 38.06 & 25.86 & 32.11 & 36.97 & 30.09 & 32.47 & 3.1 \\ 
        Parabel & 32.6 & 21.8 & 15.61 & 32.96 & 34.47 & 23.27 & 28.21 & 32.14 & 26.36 & 28.21 & 0.03 \\ 
        XT & 31.41 & 21.39 & 15.48 & 32.17 & 33.86 & 22.37 & 27.51 & 31.64 & 25.58 & 27.52 & 9.46 \\ 
        Slice & 30.43 & 20.5 & 14.84 & 31.07 & 32.76 & 23.08 & 27.74 & 31.89 & 26.11 & 28.13 & 0.08 \\ 
        PfastreXML & 32.56 & 22.25 & 16.05 & 33.62 & 35.26 & 26.81 & 30.61 & 34.24 & 29.02 & 30.67 & 0.26 \\ 
        Bonsai & 34.11 & 23.06 & 16.63 & 34.81 & 36.57 & 24.75 & 30.35 & 34.86 & 28.32 & 30.47 & 0.1 \\ 
        XR-Transformer & 38.1 & 25.57 & 18.32 & 38.89 & 40.71 & 28.86 & 34.85 & 39.59 & 32.92 & 35.21 & 35.4 \\

        \bottomrule
    \end{tabular}}

\end{table*}

\begin{table*} []
 \caption{Detailed results on long-text dataset. TT refers to training time in hours on a single Nvidia V100 GPU.}
    \label{tab:supp:results_short_text}
      \centering
      \resizebox{\linewidth}{!}{
        \begin{tabular}{@{}l|ccccccccccc@{}}
        \toprule
        \textbf{Method} & \textbf{P@1} & \textbf{P@3} & \textbf{P@5} & \textbf{N@3} & \textbf{N@5} & \textbf{PSP@1} & \textbf{PSP@3} & \textbf{PSP@5} & \textbf{PSN@3} & \textbf{PSN@5} & \textbf{TT} \\ \midrule
        
        \midrule
        \multicolumn{12}{c}{LF-Amazon-131K}\\ \midrule
        \lexus & 47.13 & 31.29& 22.35& 48.23& 50.47& 39.14& 46.07& 51.93& 44.32& 47.28 & 7.92  \\
        \renee & 48.05& 32.33& 23.26 & 49.56& 52.04 & 40.11& 47.39 & 53.67& 45.37& 48.52 & 6.97\\
        DEXA & 47.12 & 31.35 & 22.35 & 48.06 & 50.22 & 38.86 & 45.17 & 50.59 & 43.29 & 45.99 &41.41 \\
        DEXA (ensemble) & 47.16 & 31.45 & 22.42 & 48.2 & 50.36 & 38.7 & 45.43 & 50.97 & 43.44 & 46.19 & 41.41 \\
        NGAME & 46.53 & 30.89 & 22.02 & 47.44 & 49.58 & 38.53 & 44.95 & 50.45 & 43.07 & 45.81 & 39.99\\
        NGAME (ensemble) & 46.65 & 30.95 & 22.03 & 47.51 & 49.61 & 38.67 & 44.85 & 50.12 & 43 & 45.64  & 39.99\\
        SiameseXML & 44.81 & 30.19 & 21.94 & 46.15 & 48.76 & 37.56 &	43.69 & 49.75 & 41.91 & 44.97 & 1.18 \\ 
        ECLARE & 43.56 & 29.65 & 21.57 & 45.24 & 47.82 & 34.98 & 42.38 & 48.53 & 40.3 & 43.37 & 2.15 \\ 
        GalaXC & 41.46 & 28.04 & 20.25 & 43.08 & 45.32 & 35.1 & 41.18 & 46.38 & 39.55 & 42.13 & 0.45 \\ 
        DECAF & 42.94 & 28.79 & 21 & 44.25 & 46.84 & 34.52 & 41.14 & 47.33 & 39.35 & 42.48 & 1.8 \\ 
        Astec & 42.22 & 28.62 & 20.85 & 43.57 & 46.06 & 32.95 & 39.42 & 45.3 & 37.45 & 40.35 & 3.05 \\ 
        AttentionXML & 42.9 & 28.96 & 20.97 & 44.07 & 46.44 & 32.92 & 39.51 & 45.24 & 37.49 & 40.33 & 50.17 \\ 
        MACH & 34.52 & 23.39 & 17 & 35.53 & 37.51 & 25.27 & 30.71 & 35.42 & 29.02 & 31.33 & 13.91 \\ 
        LightXML & 41.49 & 28.32 & 20.75 & 42.7 & 45.23 & 30.27 & 37.71 & 44.1 & 35.2 & 38.28 & 56.03 \\ 
        BERTXML & 42.59	& 28.39 & 20.27	& 43.57 & 45.61 & 33.55 & 40.83 & 46.4 & 38.8 & 41.61 & 48.11 \\
        AnneXML & 35.73 & 25.46 & 19.41 & 37.81 & 41.08 & 23.56 & 31.97 & 39.95 & 29.07 & 33 & 0.68 \\ 
        DiSMEC & 41.68 & 28.32 & 20.58 & 43.22 & 45.69 & 31.61 & 38.96 & 45.07 & 36.97 & 40.05 & 7.12 \\ 
        Parabel & 39.57 & 26.64 & 19.26 & 40.48 & 42.61 & 28.99 & 35.36 & 40.69 & 33.36 & 35.97 & 0.1 \\ 
        XT & 34.31 & 23.27 & 16.99 & 35.18 & 37.26 & 24.35 & 29.81 & 34.7 & 27.95 & 30.34 & 1.38 \\ 
        Slice & 32.07 & 22.21 & 16.52 & 33.54 & 35.98 & 23.14 & 29.08 & 34.63 & 27.25 & 30.06 & 0.11 \\ 
        PfastreXML & 35.83 & 24.35 & 17.6 & 36.97 & 38.85 & 28.99 & 33.24 & 37.4 & 31.65 & 33.62 & 1.54 \\ 
        Bonsai & 40.23 & 27.29 & 19.87 & 41.46 & 43.84 & 29.6 & 36.52 & 42.39 & 34.43 & 37.34 & 0.4 \\ 
        XR-Transformer & 45.61 & 30.85 & 22.32 & 47.1 & 49.65 & 34.93 & 42.83 & 49.24 & 40.67 & 43.91 & 38.4 \\ 

         \midrule
        \multicolumn{12}{c}{LF-Wikipedia-500K}\\ \midrule
        \lexus & 84.88 & 65.87& 51.30& 79.45&77.44& 39.26 & 51.04&56.11 &50.05 & 54.98 & 40.36  \\
        \renee & 84.95 & 66.25& 51.68& 79.79& 77.83& 39.89& 51.77& 56.70&  50.73 & 55.57 & 180.00  \\
        DEXA & 84.92 & 65.5 & 50.51 & 79.18 & 76.8 & 42.59 & 53.93 & 58.33 & 52.92 & 57.44 & 57.51 \\ 
        DEXA (ensemble) & 83.52 & 65.02 & 50.85 & 78.4 & 76.69 & 42.15 & 51.89 & 57.38 & 51.36 & 56.46  \\
        NGAME & 84.01 & 64.69 & 49.97 & 78.25 & 75.97 & 41.25 & 52.57 & 57.04 & 51.58 & 56.11 & 54.88  \\
        NGAME (ensemble) & 84.01 & 64.69 & 49.97 & 78.25 & 75.97 & 41.25 & 52.57 & 57.04 & 51.58 & 56.11 & 54.88  \\ 
        \dexml{} & 85.78 & - & 50.53 & - & 77.11 & - & - & 58.97 & - & - & $\sim$592 \\
        SiameseXML & 67.26 & 44.82 & 33.73 & 56.64 & 54.29 & 33.95 & 35.46 & 37.07 & 36.58 & 38.93 & 4.37 \\ 
        ECLARE & 68.04 & 46.44 & 35.74 & 58.15 & 56.37 & 31.02 & 35.39 & 38.29 & 35.66 & 38.72 & 9.4 \\ 
        GalaXC & 55.26 & 35.07 & 26.13 & 45.51 & 43.7 & 31.82 & 31.26 & 32.47 & 32.75 & 34.5 & - \\ 
        DECAF & 73.96 & 54.17 & 42.43 & 66.31 & 64.81 & 32.13 & 40.13 & 44.59 & 39.57 & 43.7 & 44.23 \\ 
        Astec & 73.02 & 52.02 & 40.53 & 64.1 & 62.32 & 30.69 & 36.48 & 40.38 & 36.33 & 39.84 & 20.35 \\ 
        AttentionXML & 82.73 & 63.75 & 50.41 & 76.56 & 74.86 & 34 & 44.32 & 50.15 & 42.99 & 47.69 & 110.6 \\ 
        MACH & 52.78 & 32.39 & 23.75 & 42.05 & 39.7 & 17.65 & 18.06 & 18.66 & 19.18 & 20.45 & - \\ 
        X-Transformer & 76.95 & 58.42 & 46.14 & - & - & - & - & - & - & - &  \\ 
        LightXML & 81.59 & 61.78 & 47.64 & 74.73 & 72.23 & 31.99 & 42 & 46.53 & 40.99 & 45.18 & 185.56 \\ 
        AnneXML & 64.64 & 43.2 & 32.77 & 54.54 & 52.42 & 26.88 & 30.24 & 32.79 & 30.71 & 33.33 & 15.50 \\ 
        DiSMEC & 70.2 & 50.6 & 39.7 & 42.1 & 40.5 & 31.2 & 33.4 & 37 & 33.7 & 37.1 &  - \\ 
        Parabel & 68.7 & 49.57 & 38.64 & 60.51 & 58.62 & 26.88 & 31.96 & 35.26 & 31.73 & 34.61 & 2.72 \\ 
        XT & 64.48 & 45.84 & 35.46 & - & - & - & - & - & - & - & - \\ 
        PfastreXML & 59.5 & 40.2 & 30.7 & 30.1 & 28.7 & 29.2 & 27.6 & 27.7 & 28.7 & 28.3 & 63.59 \\ 
        Bonsai & 69.2 & 49.8 & 38.8 & 60.99 & 59.16 & 27.46 & 32.25 & 35.48 & - & - & -\\ 
        XR-Transformer & 81.62 & 61.38 & 47.85 & 74.46 & 72.43 & 33.58 & 42.97 & 47.81 & 42.21 & 46.61 & 318.9 \\

        \bottomrule
    \end{tabular}}

\end{table*}

\begin{table}[]
\caption{Detailed results on the proprietary datasets with 20M and 120M labels comparing \lexus{} with SOTA XC methods.}
\label{tab:rq1_appendix}
\resizebox{\textwidth}{!}{

    \centering
    \begin{tabular}{c|rrrrr|crrcr}
    \hline
    \multirow{2}{*}{Methods} & \multicolumn{5}{c|}{SS-20M \textbf{(N = 146M, V = 2 GPUs)}}                                                                                                                            & \multicolumn{5}{c}{SS-120M \textbf{(N = 370M, V = 8 GPUs)}}                                                                                                                                  \\ \cline{2-11} 
                             & \multicolumn{1}{l}{P@1} & \multicolumn{1}{l}{P@3} & \multicolumn{1}{l|}{P@5}            & \multicolumn{1}{c|}{TT (hrs)} & \multicolumn{1}{l|}{Speed-up} & \multicolumn{1}{l}{P@1}            & \multicolumn{1}{l}{P@3} & \multicolumn{1}{l|}{P@5}            & \multicolumn{1}{c|}{TT (hrs)} & \multicolumn{1}{l}{Slowdown} \\ \hline
    \ngame                    & 70.46                   & 52.32                   & \multicolumn{1}{r|}{43.94}          & \multicolumn{1}{r|}{295.83}   & 2.3x                          & \multicolumn{3}{c|}{\textbf{--Not-scalable--}}                                                     & \multicolumn{1}{c|}{-}        & \multicolumn{1}{c}{-}        \\
    \renee                    & \underline{ 71.32}             & \underline{ 54.33}             & \multicolumn{1}{r|}{\textbf{46.64}} & \multicolumn{1}{r|}{520.83}   & 4x                            & \multicolumn{1}{r}{\textbf{83.78}} & \textbf{55.30}          & \multicolumn{1}{r|}{\underline{ 41.27}}    & \multicolumn{1}{r|}{375.16}   & 15x                          \\
    \lexus                    & \textbf{71.62}          & \textbf{54.53}          & \multicolumn{1}{r|}{\underline{ 46.60}}    & \multicolumn{1}{r|}{130.23}   & 1x                            & \multicolumn{1}{r}{\underline{ 83.37}}    & \underline{ 55.17}             & \multicolumn{1}{r|}{\textbf{41.67}} & \multicolumn{1}{r|}{25.04}    & 1x                           \\ \hline
    \end{tabular}
}

\end{table}

\begin{table}[]
\centering
\caption{Detailed results on ablative study of negative mining strategies in \lexus{} on three datasets. In all the rows, the total number of negative labels per query is fixed (to 2K). Hard negatives obtained using label embeddings are denoted as ``$\enc$-hard''; hard negatives obtained using classifiers are denoted as ``$\w$-hard''. 
``Curriculum learning'' denotes gradually increasing the ratio of $\w$-hard : random negatives through the training epochs. \textit{All the hard negatives are stale (refreshed every 5 epochs)}. The best results are in \textbf{bold}; the second best \underline{underlined}. }
\label{tab:rq3_appendix}

\resizebox{\textwidth}{!}
{

\begin{tabular}{c|rrr|rrr|rrr}
\hline
\multirow{2}{*}{\textbf{Negative Sampling Strategy}} & \multicolumn{3}{c|}{\textbf{LF-AmazonTitles-131K}}                                                      & \multicolumn{3}{c|}{\textbf{LF-Amazon-131K}}                                                            & \multicolumn{3}{c}{\textbf{LF-Wikipedia-500K}}                                                         \\ \cline{2-10} 
                                  & \multicolumn{1}{l}{\textbf{P@1}} & \multicolumn{1}{l}{\textbf{P@3}} & \multicolumn{1}{l|}{\textbf{P@5}} & \multicolumn{1}{l}{\textbf{P@1}} & \multicolumn{1}{l}{\textbf{P@3}} & \multicolumn{1}{l|}{\textbf{P@5}} & \multicolumn{1}{l}{\textbf{P@1}} & \multicolumn{1}{l}{\textbf{P@3}} & \multicolumn{1}{l}{\textbf{P@5}} \\ \hline
random                             & 44.57                            & 30.24                            & \underline{21.83}                    & 45.62                            & 31.28                            & \textbf{22.90}                              & 72.49                            & 55.85                            & 44.58                            \\ \hline
$\enc$-hard                          & 41.98                            & 27.63                            & 19.07                             & 15.39                            & 11.34                            & 8.65                              & 57.37                            & 38.29                            & 28.87                            \\
random $+ $ $\enc$-hard                 & 45.55                            & 30.24                            & 21.55                             & 45.79                            & {30.74}                   & 22.15                             & 81.40                            & 61.98                            & 47.60                            \\ 
curriculum Learning ($\enc$ hard)               & \underline{45.84}    & \underline{30.53}        & 21.75                      & 43.81  &29.44      & 21.25       & { 74.35}    &58.30        & {46.24}        \\\hline
$\w$-hard                          & 41.74                            & 28.60                            & 20.79                             & 17.86                            & 11.96                            & 8.87                              & 52.84                            & 33.30                            & 22.92                            \\
Curriculum learning               & \underline{45.83}                      & {30.40}                      & 21.58                             & \textbf{47.33}                   & \textbf{31.50}                      & \underline{22.47}                    & \underline{84.56}                      & \textbf{66.23}                   & \textbf{51.72}                   \\
\rowcolor{Gray}
random $+ $ $\w$-hard (\lexus)           & \textbf{46.20}                   & \textbf{30.51}                   & \textbf{21.95}                       & \underline{47.13}                      & \underline{31.29}                            & {22.35}                       & \textbf{84.88}                   & \underline{66.09}                      & \underline{51.30}                      \\ \hline

\end{tabular}
}

\end{table}

\begin{table*}[]
\caption{Results (Precision@$k$) on public short-text datasets without label features comparing \lexus{} with \sota{} XC methods. Results for \lightxml, \renee{} are from \citet{renee_2023}, \dexa{} and \ngame{}  are from \citet{dexapaper}; and \cascadexml{} and \elias{} numbers are reported from \citet{buvanesh2024enhancing}. The best results are in \textbf{bold}; the second best \underline{underlined}. }
\label{tab:rq2_table_2}
\centering
\resizebox{\textwidth}{!}{
\begin{tabular}{c|rrrrrr|rrrrrr}
\hline
\multirow{2}{*}{\textbf{Methods}} & \multicolumn{6}{c|}{\textbf{Amazon-670K}}                                                                                                                                                                                 & \multicolumn{6}{c}{\textbf{Amazon-3M}}                                                                                                                                                                                   \\ \cline{2-13} 
                                  & \multicolumn{1}{l}{\textbf{P@1}} & \multicolumn{1}{l}{\textbf{P@3}} & \multicolumn{1}{l|}{\textbf{P@5}}   & \multicolumn{1}{l}{\textbf{PSP@1}} & \multicolumn{1}{l}{\textbf{PSP@3}} & \multicolumn{1}{l|}{\textbf{PSP@5}} & \multicolumn{1}{l}{\textbf{P@1}} & \multicolumn{1}{l}{\textbf{P@3}} & \multicolumn{1}{l|}{\textbf{P@5}}   & \multicolumn{1}{l}{\textbf{PSP@1}} & \multicolumn{1}{l}{\textbf{PSP@3}} & \multicolumn{1}{l}{\textbf{PSP@5}} \\ \hline
\attentionxml                      & 47.14                            & 42.7                             & \multicolumn{1}{r|}{38.99}          & 32.13                              & 35.14                              & 37.82                               & 50.86                            & 48.04                            & \multicolumn{1}{r|}{45.83}          & 15.52                              & 18.45                              & 20.6                               \\
\xrtransformer                     & 49.11                            & 43.8                             & \multicolumn{1}{r|}{40.00}          & 29.90                              & 34.35                              & 38.63                               & 52.60                            & 49.40                            & \multicolumn{1}{r|}{46.90}          & \textbf{16.54}                     & \textbf{19.94}                     & \textbf{22.39}                     \\
\lightxml                          & 47.30                            & 42.20                            & \multicolumn{1}{r|}{38.50}          & \multicolumn{1}{c}{-}              & \multicolumn{1}{c}{-}              & \multicolumn{1}{c|}{-}              & \multicolumn{6}{c}{--Not-scalable--}                                                                                                                                                                                     \\
\elias                             & \multicolumn{1}{l}{48.68}        & \multicolumn{1}{l}{43.78}        & \multicolumn{1}{l|}{40.04}          & 31.22                              & 34.94                              & 38.31                               & \multicolumn{1}{l}{49.93}        & \multicolumn{1}{l}{47.07}        & \multicolumn{1}{l|}{44.85}          & 14.97                              & 17.46                              & 19.34                              \\
\cascadexml                        & 48.80                            & 43.80                            & \multicolumn{1}{r|}{40.1}           & 31.40                              & 36.22                              & 40.28                               & 51.30                            & 49.00                            & \multicolumn{1}{c|}{46.90}          & -                                  & -                                  & -                                  \\
\renee                             & \textbf{54.23}                   & \textbf{48.22}                   & \multicolumn{1}{r|}{\textbf{43.83}} & \textbf{34.16}                     & \textbf{39.14}                     & \textbf{43.39}                      & \textbf{54.84}                   & \textbf{52.08}                   & \multicolumn{1}{r|}{\textbf{49.77}} & 15.74                              & \underline{ 19.06}                        & 21.34                              \\
\textbf{\lexus}                    & \underline{ 53.05}                      & \underline{ 47.28}                      & \multicolumn{1}{r|}{\underline{ 42.96}}    & \underline{ 33.21}                        & \underline{ 38.40}                        & \underline{ 42.78}                         & \underline{ 52.75}                      & \underline{ 49.73}                      & \multicolumn{1}{r|}{\underline{ 47.40}}    & \underline{ 15.78}                        & 18.97                              & \underline{ 21.37}                        \\ \hline
\end{tabular}
}
\end{table*}

\begin{table}[]
\caption{Comparison of dualencoder models with \lexus{}.} 
 \label{tab:dualencoder}
\centering

\begin{tabular}{l|rr|rr|rr}
\hline
\multirow{2}{*}{\textbf{Method}} &
  \multicolumn{2}{c|}{\textbf{LF-AmazonTitles-131K}} &
  \multicolumn{2}{c|}{\textbf{LF-AmazonTitles-1.3M}} &
  \multicolumn{2}{c}{\textbf{LF-Wikipedia-500K}} \\ \cline{2-7} 
               & \textbf{P@1}   & \textbf{P@5}   & \textbf{P@1}   & \textbf{P@5}   & \textbf{P@1}   & \textbf{P@5}   \\ \hline
\textbf{\dexml} & 42.52          & 20.64          & \textbf{58.40} & \textbf{45.46} & \textbf{85.78} & \underline{50.53}    \\
\textbf{\dexa}  & \underline{44.76}    & 21.18          & 51.92          & 38.86          & 79.99          & 42.52          \\
\textbf{\ngame} & 42.61          & 20.69          & 45.82          & 35.48          & 77.92          & 40.95          \\
\textbf{ANCE}  & 42.67          & 20.98          & 46.44          & 37.59          & 63.33          & 43.35          \\
\textbf{DPR}   & 41.85          & 20.88          & 44.64          & 34.83          & 65.23          & 45.85          \\
\textbf{\lexus} & \textbf{46.20} & \textbf{21.95} & \underline{55.71}    & \underline{44.26}    & \underline{84.88}    & \textbf{51.30} \\ \hline
\end{tabular}

\end{table}

\begin{table*}[]
\caption{Ablation study on the effect of augmenting training data with label text on \lexus{} accuracy. } 
 \label{tab:data_aug}
\centering

\begin{tabular}{l|rr|rr}
\hline
\multicolumn{1}{c|}{\multirow{2}{*}{\textbf{Methods}}} & \multicolumn{2}{c|}{\textbf{LF-AmazonTitles-131K}} & \multicolumn{2}{c}{\textbf{LF-Wikipedia-500K}} \\ \cline{2-5} 
\multicolumn{1}{c|}{}            & \textbf{P@1} & \textbf{P@5} & \textbf{P@1} & \textbf{P@5} \\ \hline
\lexus{}(without data augmentation) & 44.81        & 21.64        & 82.88        & 49.20        \\
\lexus{}(with data augmentation)    & 46.20        & 21.95        & 84.88        & 51.30        \\ \hline
\end{tabular}

\end{table*}

\begin{table*}[]
\centering
\caption{Results with $95\%$ confidence intervals for LF-AmazonTitles-131K and LF-Wikipedia-500K}
\label{tab:stats}
\resizebox{\textwidth}{!}
{
\begin{tabular}{ccc|ccc}
\hline
\multicolumn{3}{c|}{\textbf{LF-AmazonTitles-131K}}                                                                                                            & \multicolumn{3}{c}{\textbf{LF-Wikipedia-500K}}                                                                                                               \\ \hline
\textbf{P@1}                                       & \textbf{P@3}                                       & \textbf{P@5}                                        & \textbf{P@1}                                       & \textbf{P@3}                                       & \textbf{P@5}                                       \\ \hline
\multicolumn{1}{r}{46.15 $\pm$ 0.037} & \multicolumn{1}{r}{30.72 $\pm$ 0.095} & \multicolumn{1}{r|}{21.91 $\pm$ 0.094} & \multicolumn{1}{r}{84.88 $\pm$ 0.093} & \multicolumn{1}{r}{65.87 $\pm$ 0.076} & \multicolumn{1}{r}{51.30 $\pm$ 0.065} \\ \hline
\end{tabular}
}
\end{table*}









\end{document}